\documentclass{article}

\usepackage[preprint]{neurips_2026}

\usepackage[utf8]{inputenc} % allow utf-8 input
\usepackage[T1]{fontenc}    % use 8-bit T1 fonts
\usepackage{hyperref}       % hyperlinks
\usepackage{url}            % simple URL typesetting
\usepackage{booktabs}       % professional-quality tables
\usepackage{amsfonts}       % blackboard math symbols
\usepackage{nicefrac}       % compact symbols for 1/2, etc.
\usepackage{microtype}      % microtypography
\usepackage{xcolor}         % colors

\usepackage{amsmath, amsthm, amssymb}
\usepackage{bbm}
\usepackage{graphicx}
\usepackage{subfig}
\usepackage{enumitem}
\usepackage[most]{tcolorbox}
\tcbuselibrary{listings,breakable}
\usepackage{listings}
\usepackage{fontawesome5}
\usepackage{wrapfig}
\usepackage{float}
\usepackage{multirow}

\newcommand{\Var}{\mathrm{Var}}

\tcbset{
  base/.style={
    arc=0mm, 
    bottomtitle=0.5mm,
    boxrule=0mm,
    colbacktitle=black!10!white, 
    coltitle=black, 
    fonttitle=\bfseries, 
    left=1mm,
    leftrule=1mm,
    right=1mm,
    top=1mm,
    bottom=1mm,
    title={#1},
    toptitle=0.75mm, 
    width=\textwidth
  }
}

\newtcolorbox{greybox}[1]{
  colframe=black!30!white,
  colback=black!5!white,
  base={#1},
  breakable
}

\floatstyle{plain}
\newfloat{prompt}{htbp}{lop}
\floatname{prompt}{Prompt}

\newtcblisting{promptbox}[2][]{
  enhanced,
  listing only,              % content is treated as a listing
  colback=gray!5,            % background
  colframe=gray!80,          % frame color
  left=2mm, right=2mm, top=1mm, bottom=1mm,
  boxrule=0.5pt,
  arc=2mm,
  outer arc=2mm,
  title = {#2},
  breakable=false,
  fontupper=\small,
  listing options={
    basicstyle=\ttfamily\small,     % code font
    breaklines=true,                % crucial: wrap long lines
    breakindent=0pt,
    columns=flexible,               % better handling of spacing
    keepspaces=true,                % preserve spaces (important for indentation)
    showstringspaces=false,
    #1                              % allow passing listings keys, e.g. language=Python
  }
}

\theoremstyle{plain}
\newtheorem{theorem}{Theorem}[section]
\newtheorem{proposition}[theorem]{Proposition}

\theoremstyle{definition}

\theoremstyle{remark}

\title{Discovery of Hidden Miscalibration Regimes}

\author{%
  Katarzyna Kobalczyk\\
  University of Cambridge\\
  \texttt{knk25@cam.ac.uk} \\
  \And
  Mihaela van der Schaar \\
  University of Cambridge \\
  \texttt{mv472@cam.ac.uk} \\
}

\begin{document}

\makeatletter
\renewcommand\paragraph{\@startsection{paragraph}{4}{\z@}%
  {1ex \@plus .5ex \@minus .2ex}% space before
  {-1em}% space after; negative = run-in heading
  {\normalfont\normalsize\bfseries}}
\makeatother

\maketitle

\begin{abstract}
Calibration is commonly evaluated by comparing model confidence with its empirical
correctness, implicitly treating reliability as a function of the confidence
score alone. However, this view can hide substantial structure: models may be
systematically overconfident on some kinds of inputs and underconfident on
others, causing global reliability diagnostics to obscure localised calibration
failures. To address this, we formulate the problem of \textit{discovering hidden
miscalibration regimes} without assuming access to predefined data slices. We
define the corresponding miscalibration field and propose a diagnostic
framework for estimating it. Our approach learns a calibration-aware
representation of the input space and estimates signed local miscalibration by
kernel smoothing in the learned geometry. Across four real-world LLM benchmarks
and twelve LLMs, we find that input-dependent calibration heterogeneity is
prevalent. We further show that the discovered fields are actionable: they
support local confidence correction and reduce calibration error in
systematically miscalibrated regions where confidence-based methods such as
isotonic regression and temperature scaling are less effective.
\end{abstract}

\section{Introduction}

\paragraph{Can calibration failures have hidden structure?}
Calibration is usually evaluated by comparing a model's reported confidence
with empirical correctness. A classifier that predicts $70\%$ confidence should
be correct roughly $70\%$ of the time, and reliability diagrams operationalise
this idea by grouping predictions according to their confidence scores~\citep{guo2017calibration}. This
view is simple and widely used, but it makes a strong aggregation assumption:
reliability is treated as a function of the confidence score alone. All inputs
assigned the same confidence are implicitly treated as exchangeable for the
purpose of calibration analysis, regardless of what those inputs are.

This assumption can fail in heterogeneous domains. A model may be
systematically overconfident on one kind of input and underconfident on
another, even when both kinds of inputs receive similar confidence scores. In
that case, global reliability diagrams can hide rather than reveal calibration
failures: opposing errors cancel when aggregated by confidence. Slice-based
analysis~\citep{hebertjohnson2018multicalibration} can expose such failures when the relevant groups are known in
advance, but in complex domains the miscalibrated regions need not align with
available metadata, dataset labels, or obvious human-defined partitions. This
issue may be especially relevant for large language models (LLMs), which are
trained on broad corpora and deployed across tasks that vary in semantics,
difficulty, format, and ambiguity.

\paragraph{Discovering hidden miscalibration regimes.}
We therefore study calibration as an input-dependent discovery problem. Without
access to predefined data slices or ground-truth data-generating probabilities,
can we recover regions of the input space where the model is systematically
overconfident or underconfident? Our central premise is that calibration
failures often follow latent structure: inputs that are similar in meaning,
context, or task type may exhibit similar calibration trends. If this structure
can be revealed, we can identify hidden calibration regimes, quantify how much
miscalibration is obscured by global diagnostics, and analyse how reliability
varies across tasks or semantic domains.

\paragraph{Learning a miscalibration field.}
To formalise this discovery problem, we define an input-dependent
miscalibration field. Unlike reliability diagrams, which condition on scalar
confidence scores, this field measures signed calibration error over the input
space itself. We estimate this field by learning a representation in which
examples with similar calibration behaviour are close together. In this learned
geometry, local averaging of residuals reveals coherent regions of over- and
underconfidence. The resulting estimator is primarily a diagnostic object: it identifies where
calibration errors occur, their direction, and how strongly they vary across
the input space. The same field can also be used downstream for local
confidence correction.

\paragraph{Calibration heterogeneity in modern LLMs.}
We first validate the framework on controlled synthetic experiments where the
true miscalibration structure is known. We then use it as a diagnostic tool in
a large-scale study of LLM predictions across four real-world benchmarks and
twelve base models from different families. The synthetic experiments show
that the method recovers known hidden miscalibration structure from noisy
residuals. On LLM benchmarks, we find that input-dependent calibration
heterogeneity appears across many settings but is not uniform: some
model--dataset pairs exhibit mostly homogeneous bias, while others contain
pronounced hidden regimes of overconfidence and underconfidence. Moreover, the
degree of heterogeneity predicts when local correction is most useful: in
highly heterogeneous settings, the learned field reduces slice-level
calibration error where confidence-based post-hoc methods can remain
unreliable.

\paragraph{Contributions.}
$\blacktriangleright$~\textbf{Hidden regime discovery.}
We formulate the problem of discovering coherent regions of over- and
underconfidence without predefined slices or subgroup labels.
$\blacktriangleright$~\textbf{Miscalibration fields.}
We define an input-dependent miscalibration field that captures where signed
calibration error varies across the input space, rather than only as a
function of confidence.
$\blacktriangleright$~\textbf{Calibration-aware geometry.}
We introduce a diagnostic framework that learns a representation in which
local residual averaging exposes stable regime-level calibration structure.
$\blacktriangleright$~\textbf{LLM evidence.}
Across four benchmarks and twelve LLMs, we show that calibration
heterogeneity is present but non-uniform, varying across datasets, model
families, and scales.
$\blacktriangleright$~\textbf{Actionable local correction.}
The learned field supports targeted confidence correction, reducing
calibration error in high-error regions where confidence-based post-hoc methods
are less effective.

\begin{figure}[t]
    \centering
    \vspace{-1em}
    \includegraphics[width=\linewidth]{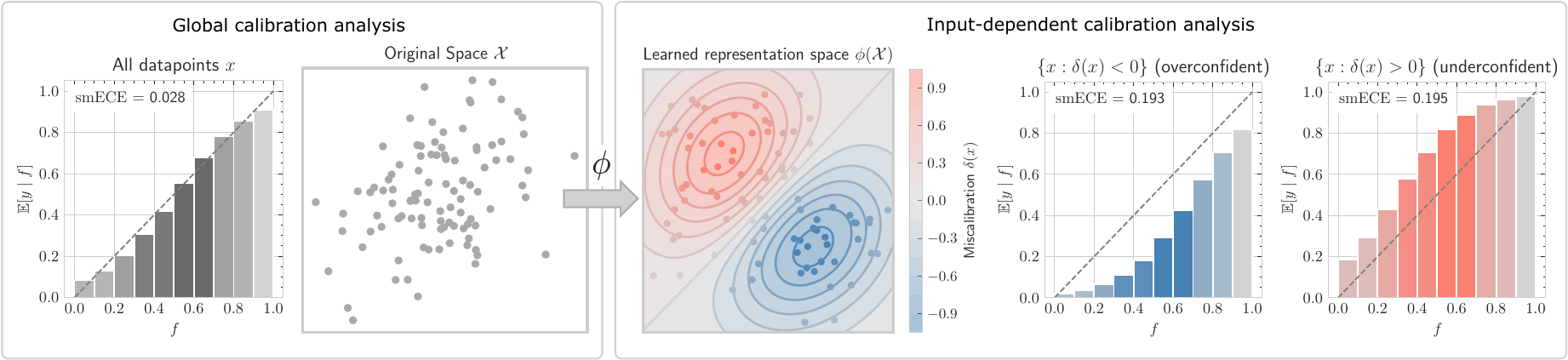}
    \caption{\textit{Hidden miscalibration regimes.}
Left: Aggregating predictions solely by confidence scores
produces a reliability diagram that suggests near-perfect calibration and low calibration error (smECE). Right: The learned representation map $\phi$ induces a geometry revealing the hidden regions of overconfidence (blue) and underconfidence (red).}
    \label{fig:figure_1}
    \vspace{-1em}
\end{figure}

\section{Background, Related Work, and Problem Setup}
\label{sec:background_problem}

In this section we introduce the notation and background, focusing on 
the notions most directly relevant to our setting; an extended discussion of related work is provided in
Appendix~\ref{app:related_work}.

\paragraph{The Classical View: Calibration in the Confidence Space.}
\label{sec:background:confidence_space}
Consider a binary classifier that assigns each input
$x\in\mathcal{X}$ a confidence score $f(x)\in[0,1]$, interpreted as an
estimate of the conditional probability
$\eta(x)=\mathbb{P}(Y=1\mid X=x)=\mathbb{E}[Y\mid X=x]$, where
$Y\in\{0,1\}$. Classical calibration asks whether predictions assigned a
given confidence are correct at the corresponding frequency
\citep{dawid1982well,degroot1983comparison}. Formally, the
calibration function is $\mu(f)=\mathbb{E}[Y\mid f(X)=f]$, and a model is
perfectly calibrated if $\mu(f)=f$ for all confidence values $f$.
Equivalently, defining the residual $R=Y-f(X)$, score-conditioned
miscalibration can be written as
$\delta(f)=\mathbb{E}[R\mid f(X)=f]$. Positive or negative values indicate
underconfidence or overconfidence, respectively.

In practice, $\mu(f)$ and $\delta(f)$ are estimated by grouping examples
with similar confidence scores. Reliability diagrams approximate the
calibration curve by binning predictions according to $f(x)$ and comparing
average confidence with empirical accuracy within each bin
\citep{degroot1983comparison,guo2017calibration}. The expected calibration
error (ECE) summarises the magnitude of these binned deviations and is widely
used in modern neural-network calibration studies \citep{guo2017calibration}.
A smooth alternative replaces hard binning with kernel smoothing in the
confidence space, yielding a smoothed reliability diagram
\citep{blasiok2024smooth} with its corresponding version of the ECE, SmoothECE (smECE). 
As we will see, our method generalises this idea: instead of
smoothing residuals among examples with similar confidence scores, we smooth
residuals among examples that are close in a learned input representation.

Post-hoc calibration methods such as Platt scaling
\citep{platt1999probabilistic}, temperature scaling
\citep{guo2017calibration}, and isotonic regression
\citep{zadrozny2001obtaining,zadrozny2002transforming} also operate in
the confidence space: they learn a mapping from the scalar score $f(x)$
to a calibrated probability. This confidence-based perspective is widely adopted, but it makes an important aggregation choice: calibration is studied as a function of the
score $f(x)$ alone. All inputs assigned similar confidence values are
treated as exchangeable for the purpose of calibration analysis, regardless
of where they lie in the input space.

\paragraph{Hidden Regimes.}\label{sec:background:hidden_regimes}
In heterogeneous datasets, the same confidence value may correspond to
different calibration behaviour depending on the type of the input $x$. A
model may be systematically overconfident on one type of inputs and
underconfident on others. When calibration is evaluated only as a function
of confidence, these effects can cancel, producing reliability diagrams that
appear well calibrated even though substantial local miscalibration remains.
Figure~\ref{fig:figure_1} illustrates this failure mode: aggregation by
confidence scores alone hides distinct regions of overconfidence and
underconfidence.

A standard way to diagnose input-dependent calibration is slice-based analysis, where
reliability is evaluated separately within predefined subgroups. This is
closely related to subgroup calibration and multicalibration, which require
calibration to hold across a collection of subpopulations
\citep{hebertjohnson2018multicalibration}. However, these approaches require the relevant slices
or subgroup classes to be specified in advance. In high-dimensional domains
such as language and vision, the regions where miscalibration occurs may not
align with dataset labels or any obvious human-defined partition.
This motivates an automatic approach: instead of asking whether a model is
calibrated on predefined groups, we aim to discover and quantify
miscalibration heterogeneity directly from observed predictions and outcomes.

\paragraph{Our Object of Study: The Miscalibration Field.}\label{sec:background:field}
We are given i.i.d.\ triples $(x_i,f_i,y_i)$, where
$x_i\in\mathcal{X}$, $f_i=f(x_i)\in[0,1]$, and $y_i\in\{0,1\}$. Our goal is
to discover regions of the input space where the model is systematically
overconfident or underconfident without assuming access to predefined data slices.

To formalise this goal, we define the input-dependent \textit{miscalibration field}
as
\begin{equation}
 \delta(x)=\mathbb{E}[Y-f(X)\mid X=x]=\mathbb{E}[R\mid X=x].   
\end{equation}
Since $\delta(x)=\eta(x)-f(x)$, the field measures the local difference
between the model's confidence and the true conditional probability. Thus
$\delta(x)<0$ indicates local overconfidence and $\delta(x)>0$ indicates
local underconfidence.

This differs from classical calibration, which studies
$\mathbb{E}[R\mid f(X)=f]$, a score-conditioned average over all inputs with
the same confidence. This distinction is key: two inputs can have the
same value of $f(x)$ while having different signs or magnitudes of
$\delta(x)$.Therefore, our objective is not to recalibrate confidence scores globally, but
to recover input-space structure indicating where and in what direction
calibration fails.

\section{A Diagnostic Framework for Revealing Calibration Regimes}
\label{sec:method}

We now describe our procedure to estimate the input-dependent miscalibration field
introduced in Section~\ref{sec:background:field}. Our goal is to
learn a representation in which examples with similar calibration behaviour
are organised close together. In such a geometry, local averaging of residuals
provides stable estimates of systematic overconfidence and underconfidence.

\subsection{Kernel Estimation in a Learned Representation}

Let $r_i := y_i - f_i$ denote the empirical residual of example $i$.
Confidence-based smoothed calibration \citep{blasiok2024smooth} estimates score-conditioned residuals by
averaging $r_i$ among examples with similar confidence values. We instead
average residuals among examples that are close in a learned representation
of the input. Let $\phi:\mathcal{X}\to\mathbb{R}^d$ be a learnable representation map.
Given a kernel bandwidth $\sigma>0$, define
\[
K_\sigma(\phi(x),\phi(x'))
:=
\exp\!\left(
-\frac{\|\phi(x)-\phi(x')\|^2}{\sigma^2}
\right).
\]
For a query point $x$, we estimate $\delta(x)$ by kernel smoothing
the residuals in $\phi(\mathcal{X})$:
\begin{equation}
\label{eq:DeltaHat}
\widehat{\delta}_\phi(x)
=
\frac{\sum_{i=1}^n K_\sigma(\phi(x),\phi(x_i))\, r_i}
     {\sum_{i=1}^n K_\sigma(\phi(x),\phi(x_i))}.
\end{equation}
This is a Nadaraya--Watson \citep{nadaraya1964study, watson1964smooth} estimator of
$\mathbb{E}[R\mid \phi(X)=\phi(x)]$, and therefore estimates a
representation-smoothed version of the miscalibration field
$\delta(x)=\mathbb{E}[R\mid X=x]$. The role of $\phi$ is to induce a
geometry in which this smoothed field remains informative, with nearby examples
sharing homogeneous calibration behaviour.

Note, kernel estimates for validation and test points are computed
using the training set as the neighbour bank. This ensures that evaluation
does not use labels from the validation or test split to construct local
neighbourhoods.

\subsection{Learning Representations that Reveal Systematic Residuals}
\label{sec:method:learning}

A useful representation should make systematic residual structure visible
under local averaging. If $\phi$ mixes examples with both positive and negative values of $\delta(x)$ within the same neighbourhood, the empirical residuals tend to cancel resulting in $\widehat{\delta}_\phi(x)$ close to zero. Conversely, if $\phi$ separates
regions with distinct calibration behaviour, local residual averages have
large magnitude and their sign reveals the corresponding over-
and underconfident regimes.

This observation motivates learning $\phi$ to maximise the variability of the
conditional mean residual, $\mathrm{Var}(\mathbb{E}[R\mid \phi(X)])$.
Since $\mathbb{E}[R]$ is independent of $\phi$, this is equivalent, up to an
additive constant, to maximising the second moment
$\mathbb{E}[(\mathbb{E}[R\mid \phi(X)])^2]$. We approximate this objective
empirically by replacing $\mathbb{E}[R\mid \phi(X)=\phi(x_i)]$ with the
kernel estimate $\widehat{\delta}_\phi(x_i)$, yielding:
\begin{equation}
\label{eq:emp_obj}
\widehat{J}(\phi)
=
\frac{1}{n}\sum_{i=1}^n \widehat{\delta}_\phi(x_i)^2.
\end{equation}
Maximizing Eq.~(\ref{eq:emp_obj}) encourages representations in which
local residual averages are non-cancelling and therefore expose systematic
overconfidence and underconfidence in the input space.

\paragraph{Preventing Neighbourhood Collapse.}
We note that the objective in Eq.~(\ref{eq:emp_obj}) admits a degenerate solution: the
representation can isolate each training point, making
$\widehat{\delta}_\phi(x_i)\approx r_i$ and artificially increasing the
objective by fitting Bernoulli noise. To prevent this, we regularise the
effective neighbourhood mass $m_i := \sum_{j=1}^n K_\sigma(\phi(x_i),\phi(x_j))$.
Our final loss for training $\phi$ is:
\begin{equation}
\label{eq:loss}
\mathcal{L}(\phi)
=
-
\frac{1}{n}\sum_{i=1}^n \widehat{\delta}_\phi(x_i)^2
+
\lambda
\frac{1}{n}\sum_{i=1}^n
\left[\max(0,m_{\min}-m_i)\right]^2,
\end{equation}
where $m_{\min}>0$ is a target minimum neighbourhood mass and
$\lambda\ge 0$ controls the strength of the regularisation. The first term
encourages discovery of systematic residual structure, while the second
discourages isolated neighbourhoods and stabilises the estimator.

\paragraph{Hyperparameter Selection.}
The training objective in Eq.~(\ref{eq:loss}) depends on the kernel
bandwidth $\sigma$ and the regularisation parameters $m_{\min}$ and $\lambda$. As a result, losses are not directly comparable
across hyperparameter settings. Moreover, in real applications the
ground-truth miscalibration field $\delta(x)$ is unknown, so oracle metrics
such as $\mathrm{Corr}(\widehat{\delta}_\phi(X),\delta(X))$ cannot be used
for model selection. To address this, we use a validation proxy based only on
the observed triples $(x_i,f_i,y_i)$. Specifically, for a candidate
representation $\phi$ we compute locally corrected predictions as
$f(x)+\widehat{\delta}_\phi(x)$ on the validation
split and evaluate the resulting Brier score. Hyperparameters are selected to minimise
this corrected validation Brier score. This additive criterion is aligned with the
fact that $\delta(x)$ is a conditional mean residual: under squared loss, the
optimal probability correction is precisely the conditional mean of
$Y-f(X)$. A detailed description of the proxy is provided in
Appendix~\ref{app:method:proxy_metric}.

\subsection{Intuition: Variance Reduction by Local Averaging}

Estimating $\delta(x)$ is difficult because individual residuals are noisy.
Under the standard model $Y\mid X=x\sim \mathrm{Bernoulli}(\eta(x))$, residuals
$R=Y-f(X)$ satisfy $\mathbb{E}[R\mid X=x]=\delta(x)$ and have conditional
variance $\mathrm{Var}(R\mid X=x)=\eta(x)(1-\eta(x))$. Thus, when $\eta(x)$ is
not close to $0$ or $1$, each residual is a high-variance observation whose
conditional mean is the local miscalibration field $\delta(x)$.
Kernel smoothing reduces this variance by averaging residuals over local
neighbourhoods. The following result formalises this variance-reduction effect
for a fixed representation:

\begin{greybox}{}
\begin{proposition}[Variance reduction]
\label{prop:variance}
Fix a representation $\phi$ and assume the kernel $K_\sigma$ is
non-negative and bounded. Then, for any query point $x$, conditional on the
inputs $\{x_i\}_{i=1}^n$,
\[
\Var\!\left(\widehat{\delta}_\phi(x)\mid \{x_i\}_{i=1}^n\right)
\le
\frac{C}{m(x)},
\]
where $m(x)=\sum_i K_\sigma(\phi(x),\phi(x_i))$ and $C>0$ depends only on the bounded
conditional variance of $R$ and the kernel. Proof: see Appendix~\ref{app:variance}.
\end{proposition}
\end{greybox}
While this result is not a recovery guarantee for the learned geometry, it explains the
variance-reduction mechanism of local averaging once $\phi$ is fixed. It also
motivates the neighbourhood-mass regulariser in Eq.~\eqref{eq:loss}: isolated
points lead to high-variance residual estimates; sufficiently large
neighbourhoods provide stable local averages. This contrasts with the alternative possibility of direct residual regression, which attempts to learn a pointwise mapping from high-variance Bernoulli residuals (see Appendix~\ref{app:exp:resreg} for an
empirical comparison).

\subsection{Actionable Use of the Learned Field: Local Recalibration}
\label{sec:method:correction}

The learned field $\widehat{\delta}_\phi(x)$ is primarily a diagnostic object:
it identifies where the model is locally overconfident or underconfident.
However, it can also be used downstream to correct model confidences.

A naive additive correction $f(x)+\widehat{\delta}_\phi(x)$ may fall outside
the valid probability range. For a confidence score $f(x)$, any valid
additive correction must lie in $[-f(x),1-f(x)]$. We therefore use a
range-aware correction map that preserves the sign of
$\widehat{\delta}_\phi(x)$ and is monotone in its magnitude, while smoothly
saturating at the feasible endpoints. For a scale parameter $\alpha>0$, define
\[
\Delta_\alpha(x)
=
\begin{cases}
- f(x)\tanh\!\left(\alpha|\widehat{\delta}_\phi(x)|\right),
& \widehat{\delta}_\phi(x)<0,\\[4pt]
(1-f(x))\tanh\!\left(\alpha|\widehat{\delta}_\phi(x)|\right),
& \widehat{\delta}_\phi(x)\ge 0.
\end{cases}
\]
The corrected confidence is then
\begin{equation}
\label{eq:range_aware_correction}
\tilde f(x)=f(x)+\Delta_\alpha(x).
\end{equation}
This guarantees $\tilde f(x)\in[0,1]$ without hard clipping. The scalar
$\alpha$ is selected on the validation split after the representation map $\phi$ has been fixed. Thus, learning the field and converting it into final
probabilities are separated: the representation is selected to reveal
local calibration structure, while the range-aware map provides a
post-hoc correction of the probabilistic predictions.

In the next section, we evaluate the proposed framework through two complementary sets of experiments: controlled synthetic datasets that allow us to study recovery of known miscalibration structures (Section~\ref{sec:exp:synth}), and real-world experiments on large language model predictions that demonstrate the presence of input-dependent calibration heterogeneity in practice (Section~\ref{sec:exp:real}).

\section{Experiments: Synthetic Illustrations}\label{sec:exp:synth}

We first use controlled synthetic experiments to validate three aspects of
the proposed estimator: (i) it can uncover hidden calibration regimes that
are invisible to score-based diagnostics, (ii) it can recover smoothly
varying miscalibration fields from noisy Bernoulli residuals, and (iii) its
hyperparameters can be selected without oracle access to the true field
$\delta(x)$. Full data-generating details and additional studies are provided in Appendix~\ref{app:exp}.

\begin{figure}[h]
    \vspace{-1em}
    \centering
    \subfloat[True and estimated clustered field.]{
        \hspace{-1.5em}
        \includegraphics[height=2.4cm]{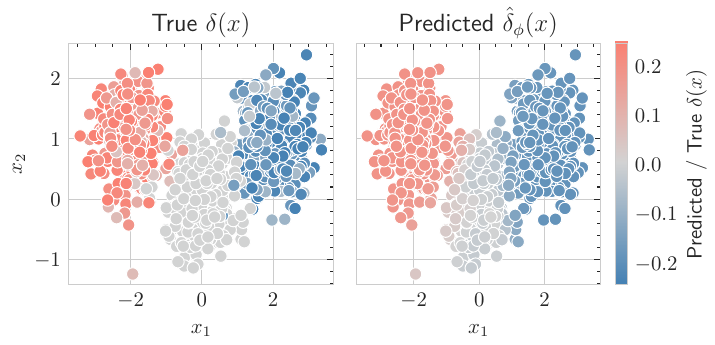}
        \label{fig:cluster_scatter_main}
    }
    \subfloat[Global calibration hides cluster-level errors.]{
        \includegraphics[height=2.4cm]{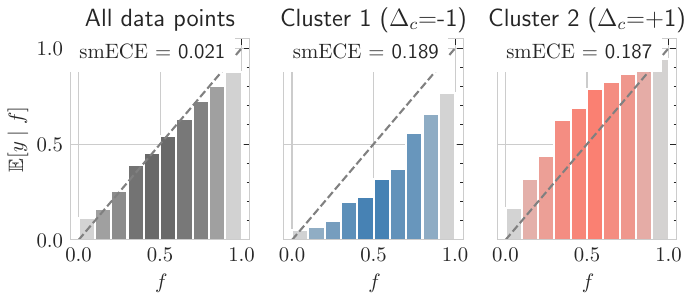}
        \label{fig:cluster_relplot_main}
    }
    \subfloat[Smooth-field recovery.]{
        \includegraphics[height=2.4cm]{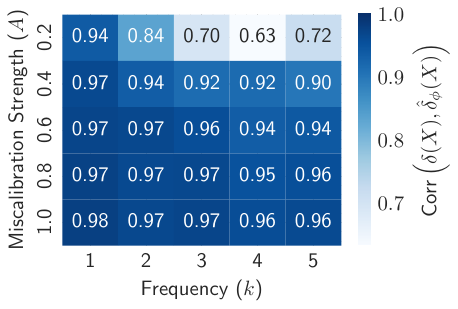}
        \label{fig:sinusoidal_pcorr_main}
    }
    \caption{
    (a, b) Clustered regimes show that global reliability can hide opposing local errors; (c) sinusoidal fields
    test recovery of smooth input-dependent miscalibration.
    }
    \vspace{-1em}
    \label{fig:synth_main}
\end{figure}

\paragraph{Setup.}
We generate two datasets: three-cluster and sinusoidal, each with $n=10{,}000$ samples constructed as triples $(x_i, f_i, y_i)$, with $x_i \in \mathbb{R}^2$.
Datasets are split into train/validation/test sets with proportions $80/10/10$. 
The representation map $\phi$ is implemented as a two-layer MLP. At validation and test
time, $\widehat{\delta}_\phi(x)$ is estimated by kernel smoothing in the
learned representation space using the training set as the neighbour bank.
Since $\delta(x)$ is known in these controlled settings, we evaluate recovery
using the Pearson correlation between $\widehat{\delta}_\phi(X)$ and
$\delta(X)$ on the test set.

\paragraph{Hidden clustered regimes.}
In the three-cluster dataset, model confidence $f(x)$ is generated from a base logit function $\ell(x)$, while the true conditional
probability includes a cluster-dependent shift. Specifically,
for a point $x$ in cluster $c(x)$, we set
$f(x)=\mathrm{sigm}(\ell(x))$ and
$\eta(x)=\mathrm{sigm}(\ell(x)+\Delta_{c(x)})$, with
$\Delta_c\in\{-1,0,+1\}$. Labels are then sampled from
$\mathrm{Bernoulli}(\eta(x))$, defining the ground-truth
miscalibration field as $\delta(x)=\eta(x)-f(x)$. This construction
induces piecewise-structured miscalibration in the
input space (Fig.~\ref{fig:cluster_scatter_main}, left). As shown in Fig.~\ref{fig:cluster_relplot_main}, the global reliability
diagram suggests near calibration because opposing errors cancel when
examples are aggregated by confidence alone. However, the cluster-level
reliability diagrams reveal substantial systematic miscalibration. The
learned field $\widehat{\delta}_\phi(x)$ recovers this hidden structure
from $(x,f,y)$ alone, without access to $\eta(x)$ or cluster labels,
achieving test correlation $r=0.95$ with the true field (Fig.~\ref{fig:cluster_scatter_main}, right).

\paragraph{Smooth miscalibration fields.}
We next replace the discrete cluster shifts with a smooth input-dependent
logit shift. As before, the model confidence is
$f(x)=\mathrm{sigm}(\ell(x))$, but the true conditional probability is
$\eta(x)=\mathrm{sigm}(\ell(x)+\Delta(x))$, where
$\Delta(x)=A\sin(2\pi k u^\top x)$. Here $u$ is a fixed unit direction,
$A$ controls the strength of miscalibration, and $k$ controls the spatial
frequency of regime alternation. This induces a continuously varying
ground-truth field $\delta(x)=\eta(x)-f(x)$ (see Appendix~\ref{app:exp:sinusoid_results} for a visualization) and tests whether the method can
recover input-dependent miscalibration beyond discrete clusters.
Fig.~\ref{fig:sinusoidal_pcorr_main} shows that the learned estimator captures the sign
and spatial structure of the true field. Across amplitudes and frequencies,
recovery remains strong and degrades mainly when the field is both weak and rapidly oscillating, that is when the signal-to-noise ratio of the residuals is lowest.

\begin{wrapfigure}{r}{0.45\linewidth}
\vspace{-3.1em}
\centering
\includegraphics[width=\linewidth]{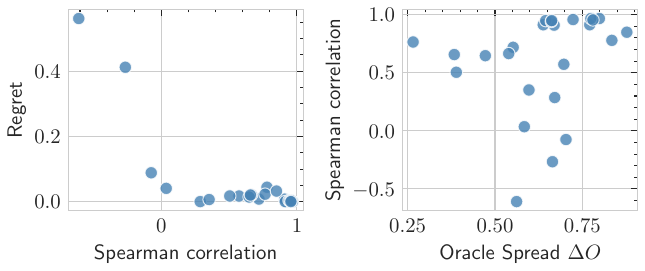}
\caption{Proxy ranking diagnostics. Left: Spearman correlation vs.\ regret.
Right: oracle spread vs.\ Spearman correlation.}
\label{fig:proxy_selection_scatter}    
\vspace{-2em}
\end{wrapfigure}
\paragraph{Proxy-based hyperparameter selection.}
The previous evaluations use the known field $\delta(x)$ only for reporting
test recovery. In practice, however, $\delta(x)$ is unobserved, so
hyperparameters such as the bandwidth $\sigma$ and mass-regularisation
strength $\lambda$ must be chosen from the observed triples
$(x_i,f_i,y_i)$ alone. In this experiment, we evaluate the validation proxy
introduced in Section~\ref{sec:method:learning}. For each sinusoidal setting,
specified by a signal strength $A$ and frequency $k$, we evaluate a grid
of hyperparameters $(\sigma,\lambda)$ and compare the proxy-selected choice
$(\hat{\sigma},\hat{\lambda})$ against the oracle objective
$O(\sigma,\lambda)=\mathrm{Corr}(\widehat{\delta}_\phi(X),\delta(X))$
computed on the test set. We summarise each setting by three quantities:
1)~the Spearman rank correlation between the negated validation proxy and
$O(\sigma,\lambda)$ across the grid, so that larger Spearman rank values indicate better
rank agreement; 2)~the oracle spread $\Delta O=O_{\max}-O_{\min}$, which
measures how much performance varies over the grid; 3)~the selection
regret $O^*-O(\hat{\sigma},\hat{\lambda})$, where
$O^*=\max_{\sigma,\lambda}O(\sigma,\lambda)$. Each point in
Fig.~\ref{fig:proxy_selection_scatter} corresponds to one sinusoidal setting
$(A,k)$. The left panel shows that stronger proxy--oracle rank agreement
is associated with lower regret. The right panel shows how rank agreement
varies with oracle spread: when the oracle surface is flat, rankings are
less stable because many hyperparameter choices have similar oracle
performance. Overall, proxy selection achieves low regret in most settings,
with median regret of just $0.008$; the full regret table by $(A,k)$
is given in Appendix~\ref{app:exp:proxy_selection}.

\begin{greybox}{}
\faLightbulb[regular]  \hspace{0.3em}  
Overall, these synthetic experiments show that learning a calibration-aware
geometry makes local residual averaging informative: it uncovers hidden
regimes that score-based diagnostics miss, recovers smooth miscalibration
fields from noisy observations, and supports practical hyperparameter
selection without oracle access to $\delta(x)$.
\end{greybox}

\paragraph{Additional experiments.}
Appendix~\ref{app:exp} provides full synthetic details and extended results:
the complete data-generating processes, sensitivity to bandwidth $\sigma$
and mass regularisation $\lambda$, extended proxy-selection diagnostics,
and a comparison with direct residual regression,
which is demonstrated to be less stable in low-signal or high-frequency regimes.

\section{Experiments: Discovering Hidden Calibration Regimes of Modern LLMs}
\label{sec:exp:real}

We next investigate whether input-dependent calibration heterogeneity arises
in real-world language model predictions. We evaluate four datasets:
HH-RLHF \citep{bai2022training}, MedMCQA \citep{flores2022medmcqa}, MMLU \citep{hendryckstest2021}, and MMLU-Pro \citep{wang2024mmlupro}; and four model families:
Qwen-3, Ministral-3, Llama-3.1/2, and Gemma-3; across varying model sizes (12 LLMs in total). 
We focus on base LLMs---prior work suggests that post-training and alignment
can degrade calibration, while base models often remain better calibrated
\citep{zhu2023calibration,xiao2025restoring,nakkiran2026trained}. Compared
with instruction- and RL-tuned models, base LLMs tend to produce confidence
scores spanning a broader range of values in $[0,1]$, making them a natural
setting for visualising and studying input-dependent calibration heterogeneity. For each dataset--model pair, we learn the representation map $\phi$ on the training
split, select hyperparameters on the validation split using our proxy defined in Section~\ref{sec:method:learning}, and report all results on the test split, using the training points as the neighbour bank. 

We organize the results around three axes: \textit{discovery}, whether the learned field identifies hidden high-error regimes; \textit{validity}, whether these regimes are meaningful and stable; and \textit{actionability}, whether the discovered structure can be used for local correction.

\paragraph{A representative example.}
We start with a representative example of the HH-RLHF dataset on Qwen3-8B. HH-RLHF consists of prompts paired with two candidate responses and a human preference label. We treat the model's
predicted preference probability for the first response as the confidence
score $f(x)$, and define $y\in\{0,1\}$ to indicate whether the first
response is preferred by the human annotator (response order is sampled randomly).

\begin{figure}[h]
\vspace{-0.25em}
\centering
\includegraphics[width=0.9\linewidth]{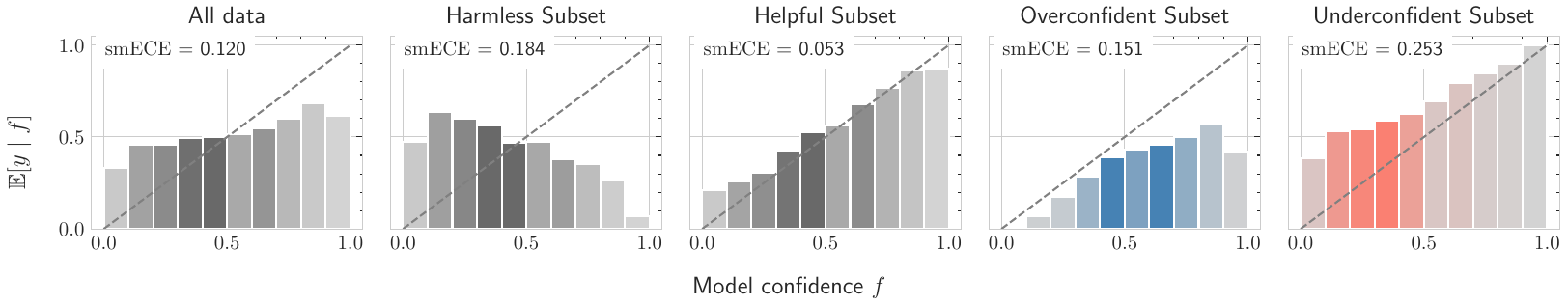}
\vspace{-0.5em}
\caption{\textit{Discovered calibration regimes on HH-RLHF and Qwen3-8B}.
Although the aggregate reliability diagram suggests only moderate
miscalibration, conditioning on $\hat{\delta}_\phi$ (last two panes) reveals
substantially stronger and opposing calibration errors across input regions.
The discovered over- and underconfident subsets expose structure that is not
captured by the original dataset slices alone.} 
\label{fig:hhrlhf_relplots_orig}
\vspace{-0.25em}
\end{figure} 

Figure~\ref{fig:hhrlhf_relplots_orig} shows reliability diagrams for the
full test set, the original HH-RLHF helpful/harmless slices, and the
regions discovered by the learned miscalibration field defined as: \textit{overconfident} $\{x : \widehat{\delta}_\phi(x) < -\epsilon\}$ and 
\textit{underconfident} $\{x : \widehat{\delta}_\phi(x) > \epsilon\}$, where we set $\epsilon=0.05$.
Although the aggregate reliability diagram suggests only moderate miscalibration,
conditioning on $\widehat{\delta}_\phi$ reveals substantially stronger
and opposing calibration errors. Importantly, this example shows that inputs with similar
confidence values can belong to different calibration regimes, and
therefore that calibration cannot be fully characterised as a function of
the scalar confidence score alone.

\begin{figure}[t]
    \centering
    \vspace{-1em}
    \subfloat[Distributional structure of the learned fields, represented by the mean and variation of $\hat{\delta}_\phi$. Negative mean values indicate overconfidence, the vertical axis captures the degree of heterogeneity. ]{
        \includegraphics[height=3.6cm]{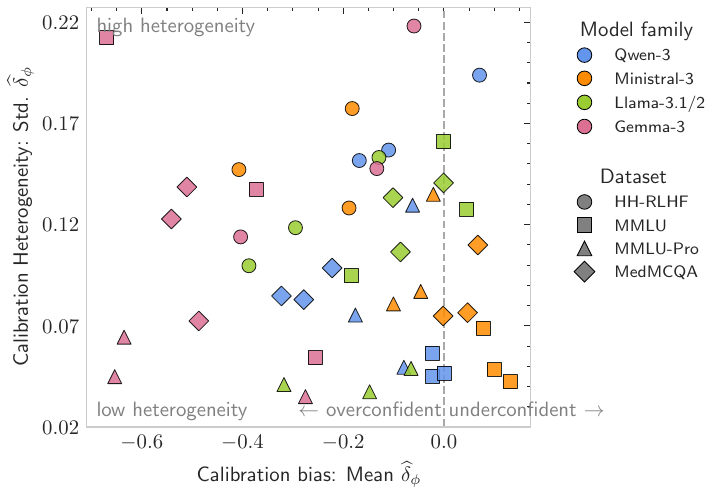}
        \quad
        \label{fig:miscalibration_map}
    }
    \quad
    \subfloat[Miscalibration heterogeneity, measured as the gap between the discovered worst-slice and global smECE on held-out test sets. Non-zero values indicate that global calibration metrics underestimate the magnitude of miscalibration in specific regions of the input space.]{
        \hspace{-2em}
        \includegraphics[height=3.4cm]{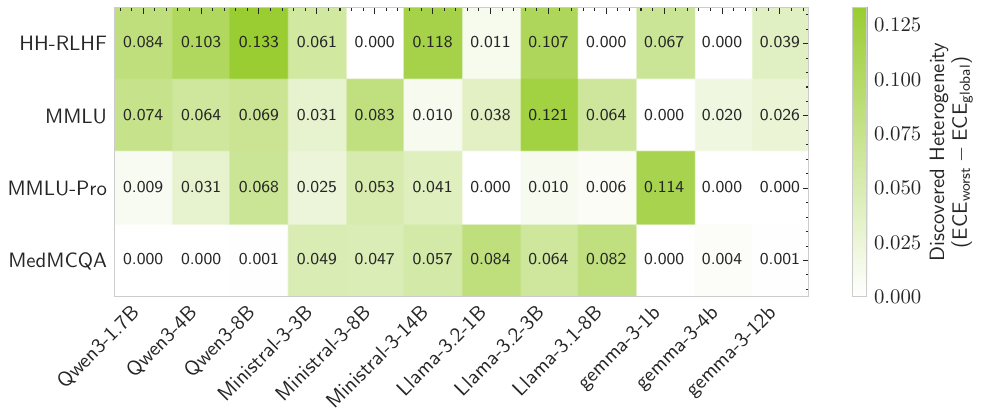}
        \label{fig:validation}
    }
    \caption{
    The learned fields exhibit structured variability (high heterogeneity points in \ref{fig:miscalibration_map}). Sign and magnitude of $\hat{\delta}_\phi$ successfully identify regions of substantial calibration error on the test sets (\ref{fig:validation}).
    }
    \label{fig:miscalibration_heterogeneity}
\vspace{-1em}
\end{figure}

\paragraph{Discovery: the learned field identifies hidden calibration regimes.}
We next characterise the learned miscalibration fields across all 48
dataset--model pairs included in this study. Figure~\ref{fig:miscalibration_heterogeneity}
separates two properties of the field: its mean, which captures the
overall calibration bias, and its standard deviation, which captures
input-dependent heterogeneity. The structure of the learned fields varies. Some settings exhibit mostly
homogeneous overconfidence, while others display high calibration heterogeneity (e.g. Qwen-3 on HH-RLHF, Llama-3.1/2 on MMLU).
Appendix~\ref{app:exp_real:characterisation} shows the same figure with points annotated by exact model sizes.
Importantly, this learned variability is meaningful. Figure~\ref{fig:validation} reports the gap between the worst-slice and global smECE on the test sets, with slices defined by $\hat{\delta}_\phi$, as previously.
We define the worst slice as the higher-smECE region among the discovered
overconfident and underconfident slices.
Large values indicate that the aggregate reliability metrics substantially
underestimate calibration error in some region of the input space. Thus,
the learned field identifies regions with genuinely worse calibration.

\paragraph{Validity: the discovered regimes are stable.}
The discovered regimes are not rare edge cases. Among the 18 dataset--model pairs
with a worst-slice smECE gap of at least $0.06$, the worst-calibrated slice
contains on average $41\%$ of test examples. Additional analyses
support the reliability of these regimes: 
Appendix~\ref{app:exp_real:robust} reruns the full pipeline under label
permutation and across random seeds. Appendix~\ref{app:bootstrap_ci} reports bootstrap
confidence intervals for worst-slice gaps and slice sizes, while
Appendix~\ref{app:threshold_sensitivity} shows robustness across thresholds. Finally, Appendix~\ref{app:exp_real:raw_vs_learned} shows that raw hidden-state smoothing mostly recovers global overconfidence, rather than signed hidden regimes.

\begin{figure}[h]
\vspace{-0.5em}
    \centering
    \includegraphics[width=\linewidth]{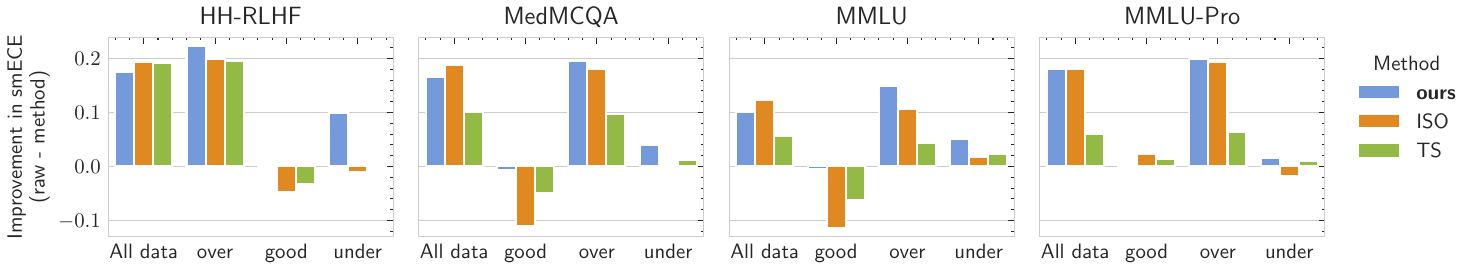}
    \caption{
\textit{Improvement in smECE relative to the base model}, aggregated across LLMs. Positive values indicate improved calibration (lower smECE). Confidence-based methods (ISO and TS) improve on the global scale (all data). Our method consistently reduces calibration error across all miscalibrated regions (over and under) while not degrading performance on well-calibrated data (good).
}
\vspace{-1em}
\label{fig:ece_improvement}
\end{figure}

\paragraph{Actionability: local correction.}
The learned field can also be used to locally correct model confidence as introduced in Section~\ref{sec:method:correction}. 
We use two confidence-based calibrators as controls for whether the discovered
structure can be explained by the scalar confidence alone: isotonic
regression (ISO) and temperature scaling (TS). Figure~\ref{fig:ece_improvement} reports the absolute improvement
in smECE relative to the raw model, aggregated across all 12 LLMs.
Confidence-based methods improve global calibration (all data), but their
behaviour is less consistent across the discovered regimes and often degrades calibration on the already well-calibrated subsets. 
In contrast, our method consistently reduces calibration error in the over- and
underconfident regions while preserving performance on the well-calibrated
subset. This demonstrates that the learned field is not only
diagnostic, but also actionable: it enables local corrections that cannot
be achieved from confidence scores alone.

\begin{figure}[t]
    \centering
    \subfloat[\textit{Variation in $\widehat\delta_\phi$ predicts when local calibration is most effective.} Points correspond to dataset--model pairs, with effectiveness defined as the improvement with respect to ISO on the worst slice.]{
        \qquad
        \includegraphics[height=3.6cm]{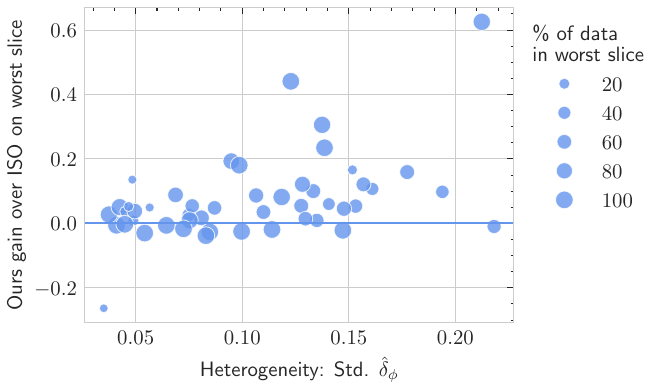}
        \quad
        \label{fig:heterogeneity_vs_gain}
    }
    \quad
    \subfloat[Heterogeneity vs. model size. In several model families, heterogeneity
increases with model size.]{
        \quad
        \includegraphics[height=3.6cm]{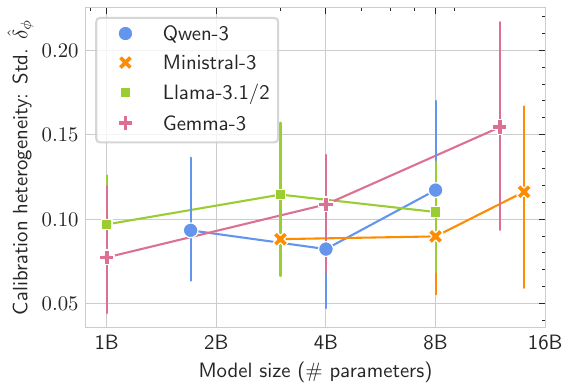}
        \quad
        \label{fig:model_size_vs_heterogeneity}
    }
    \caption{\textit{Relationship between miscalibration heterogeneity, calibration gains,
and model scale.}}
    \label{fig:relationships}
\end{figure}

\paragraph{When does local calibration help?}
We also examine when input-dependent calibration provides the largest
advantage. Figure~\ref{fig:heterogeneity_vs_gain} shows that the benefit
of our method over ISO on the worst-calibrated slice increases with the
standard deviation of the learned miscalibration field. This relationship
supports the central claim of the paper: local calibration is most useful
when calibration behaviour varies substantially across inputs.

\paragraph{What is the effect of model size?}
Finally, we study how heterogeneity varies with model scale. As shown in
Figure~\ref{fig:model_size_vs_heterogeneity}, in several model families, 
calibration heterogeneity increases with model size. One possible explanation is that model capabilities improve
non-uniformly across domains as scale increases, producing regions where
confidence becomes better aligned with correctness and others where
systematic miscalibration persists or becomes amplified. 

\begin{greybox}{}
\faLightbulb[regular]  \hspace{0.3em}  
Overall, these results suggest that input-dependent calibration is a real
phenomenon pertinent to modern LLMs. The learned fields reveal hidden
over- and underconfident regions that global diagnostics miss, and these
regions can be used for effective local correction. Gains are largest when
the learned field is most heterogeneous, supporting calibration analysis as
an input-dependent problem rather than a function of confidence alone.
\end{greybox}

\section{Discussion}

We formulated the problem of discovering hidden calibration regimes:
regions of coherent over- or underconfidence that can be obscured by
confidence-based diagnostics. To study this failure mode, we define a
miscalibration field over the input space and estimate it by learning a
calibration-aware geometry in which local averaging of residuals reveals
coherent regime-level structure. Synthetic experiments validate recovery of
known miscalibration fields, while LLM experiments show substantial
input-dependent calibration heterogeneity across datasets and model families.

The proposed estimator should be viewed as a diagnostic instrument rather than
a complete solution to local calibration. Its purpose is to reveal signed
residual structure and to test whether the learned field is actionable. The local
recalibration results illustrate that the discovered regimes can support
targeted confidence correction beyond confidence-only transformations, while
richer input-dependent calibrators remain an important complementary direction.
The current formulation focuses on binary correctness, leaving multiclass
settings and open-ended generation for future work. Finally, the learned
geometry provides a starting point for interpretability (see Appendix~\ref{app:interpretability}): although this work does not explain why a regime is miscalibrated, the representation
can be inspected using metadata or semantic descriptors
opening a path toward interpretable calibration debugging.

Overall, our results suggest that calibration should not be viewed only as a
global property of confidence scores. Discovering hidden over- and
underconfident regions is an important step toward more reliable uncertainty
estimates for models deployed across heterogeneous tasks and domains.

\clearpage

\bibliography{references}
\bibliographystyle{abbrv}

\clearpage
\appendix
{\Large \textbf{Appendix}}

\paragraph{Appendix roadmap.}
The appendix is organised as follows:
\begin{itemize}[left=1em, itemsep=0pt, topsep=2pt]
    \item \textbf{Appendix~\ref{app:related_work}} expands the related-work discussion and clarifies the distinction between this work and: confidence-based calibration, multicalibration, error-slice discovery and related works on uncertainty estimation in LLMs.
    \item \textbf{Appendix~\ref{app:method:proxy_metric}} describes in detail the validation proxy used for hyperparameter selection.
    \item \textbf{Appendix~\ref{app:variance}} proves the variance-reduction result for kernel smoothing in a learned representation.
    \item \textbf{Appendix~\ref{app:implementation}} provides implementation details, including model architecture, optimisation, datasets, prompts, and calibration baselines.
    \item \textbf{Appendix~\ref{app:exp}} gives extended synthetic results, including data generation, smooth-field recovery, hyperparameter interactions, proxy-selection diagnostics, and the residual-regression baseline.
    \item \textbf{Appendix~\ref{app:exp_real}} reports additional LLM results, including annotated field characterisation, raw-versus-learned representation comparisons, robustness audits, threshold sensitivity, bootstrap confidence intervals, post-hoc interpretability analysis, and detailed reliability plots per each model--dataset pair.
\end{itemize}

% To facilitate reproducibility, we provide an anonymised repository containing the code, experiment configurations, and instructions required to reproduce the reported results: \url{https://anonymous.4open.science/r/miscalibration-discovery-0974}.

\section{Extended Related Work}\label{app:related_work}

Our work sits at the intersection of calibration, subgroup reliability, and
automatic slice discovery, but differs from each in its object of study. Standard
post-hoc calibration methods estimate reliability as a function of the scalar
confidence score. Subgroup calibration and multicalibration ask whether
calibration holds over a specified class of groups or auditors. Slice-discovery
methods search for coherent regions of high error or degraded performance that
can be measured directly, for example through accuracy or loss. In contrast, we
formulate the discovery of \emph{signed input-dependent miscalibration} as the
central problem: given only observed triples $(x_i,f_i,y_i)$, recover regions
where the conditional residual $Y-f(X)$ is systematically positive or negative.
This defines a miscalibration field over the input space, rather than a scalar
calibration curve, a fixed subgroup constraint, or an unsigned error slice. The
representation-learning procedure we propose is designed specifically to
estimate this field and expose regimes of over- and underconfidence that can
cancel under confidence-conditioned diagnostics.

\paragraph{Confidence-based calibration.}
Classical calibration studies the relationship between a model's predicted
confidence and its empirical correctness. Early work on probabilistic
forecasting formalised calibration as a property of conditional outcome
frequencies given predicted probabilities, and reliability diagrams became a
standard tool for visualising this relationship
\citep{dawid1982well,degroot1983comparison}. Calibration is commonly evaluated
by binning predictions by confidence and comparing average confidence with
empirical accuracy, alongside scalar metrics such as expected calibration error
(ECE) \citep{guo2017calibration}. A related line of work studies post-hoc
recalibration methods that learn a one-dimensional map from model scores to
calibrated probabilities, including Platt scaling
\citep{platt1999probabilistic}, isotonic regression
\citep{zadrozny2002transforming}, and temperature scaling
\citep{guo2017calibration}. More recent work replaces hard binning with smooth
nonparametric estimates in confidence space, leading to smoothed reliability
diagrams and SmoothECE \citep{blasiok2024smooth}. These methods provide
effective score-level diagnostics and recalibration, but they do not
distinguish examples with identical confidence scores that belong to different
calibration regimes.

\paragraph{Slice analysis, subgroup calibration, and multicalibration.}
A complementary line of work argues that aggregate calibration can mask
systematic failures on subpopulations. In fairness and risk prediction,
calibration has been studied across protected groups, revealing both the
importance of group-conditional reliability and fundamental tensions between
calibration and other fairness criteria
\citep{kleinberg2017inherent,chouldechova2017fair,obermeyer2019dissecting}.
More generally, subgroup calibration and multicalibration require calibration
to hold not only globally, but simultaneously over a collection of
subpopulations or tests
\citep{hebertjohnson2018multicalibration,hansen2024when}. These frameworks
formalise the idea that reliability should be assessed beyond the aggregate
population and provide strong guarantees when the relevant subgroup class is
specified. However, they typically rely on a predefined family of groups,
auditors, or test functions. Our setting is complementary: rather than requiring
calibration over a predefined group family or auditor class, we aim to discover
the calibration-relevant geometry itself.

\paragraph{Automatic subgroup and error-slice discovery.}
Automatic slice-discovery methods aim to find coherent subsets on which a model
underperforms. \cite{chung2020automated} and \cite{sagadeeva2021sliceline}
identify interpretable high-error subpopulations for model validation and
debugging. Representation-based approaches extend this idea to higher-dimensional
settings; \cite{eyuboglu2022domino}, for example, combines cross-modal
embeddings with an error-aware mixture model to identify and describe error
slices. In NLP and LLM settings, \cite{hua2023discover} introduce an automatic
slice-detection benchmark for NLP, \cite{otey2025representing} study
representation choices for clustering offensive-language errors, and
\cite{zhang2025active} formalise active slice discovery for LLMs with limited
slice-membership annotation. Unlike error-slice discovery, which typically
targets elevated loss or misclassification rate, our target is signed residual
structure: a region may have the same accuracy as another while differing in
whether the model is overconfident or underconfident.

\paragraph{Calibration studies of LLMs.}
Recent work studies calibration in large language models across several notions
of confidence, including token probabilities, multiple-choice answer
probabilities, verbalised confidence, and semantic uncertainty.
\cite{zhu2023calibration} provide a systematic study across pretraining and
alignment stages, showing that calibration depends strongly on training stage,
task format, and alignment procedure. A recurring finding is that pretrained or
base models can exhibit meaningful calibration, while fine-tuning, instruction
tuning, preference alignment, and chain-of-thought reasoning can degrade it. For
example, \cite{he2025preserving} show that fine-tuned pretrained language
models can become overconfident, especially under distribution shift;
\cite{xiao2025restoring} study calibration degradation after preference
alignment and propose calibration-aware fine-tuning methods; and
\cite{nakkiran2026trained} show that base LLMs can be well calibrated over
semantic answer classes, while RL instruction tuning and chain-of-thought can
break this calibration. Other work develops methods for eliciting or improving
LLM confidence estimates, including calibration tuning
\citep{kapoor2024calibration}. Our work is complementary: rather than asking
only whether an LLM is globally calibrated, we show that LLM embeddings encode
information that can be used to learn input-dependent miscalibration fields,
revealing hidden regions of overconfidence and underconfidence missed by
aggregate diagnostics.

\paragraph{Broader impact statement.}
This work aims to improve model reliability by identifying input regions where
confidence is systematically miscalibrated. Such diagnostics can
support safer auditing, model debugging, and decisions about when to defer or
seek additional validation. However, the method should not be interpreted as a
guarantee of safety or correctness. Discovered regimes depend on the available
labelled data and may miss failures on underrepresented groups or deployment
conditions. If used uncritically, local confidence correction could create a
false sense of reliability, especially in high-stakes settings. We therefore
view the method as one component of a broader evaluation pipeline, to be used
alongside task-specific validation, fairness analysis, and human oversight.

\section{Proxy metric for hyperparameter selection.}
\label{app:method:proxy_metric}

In practice, the ground-truth miscalibration field $\delta(x)$ is
unknown, and oracle metrics such as
$\mathrm{Corr}(\widehat{\delta}_\phi(X), \delta(X))$ cannot be used for
model selection. Moreover, the training objective in
Eq.~(\ref{eq:loss}) depends explicitly on the kernel bandwidth $\sigma$
and the representation $\phi$, and is therefore not directly comparable
across different hyperparameter settings. In particular, a lower
validation loss does not necessarily imply better recovery of the
miscalibration field. We therefore introduce a proxy criterion for
hyperparameter selection that relies only on observable quantities
$(x_i, f_i, y_i)$ and the estimated local correction
$\widehat{\delta}_\phi(x)$.

The key idea is to interpret $\widehat{\delta}_\phi(x)$ as a local
probability correction to the base confidence $f(x)$. For a given input,
we define the corrected prediction
\[
\tilde{f}_{\text{raw}}(x)
:=
f(x) + \widehat{\delta}_\phi(x).
\]

We evaluate the quality of this correction on a validation set using the
Brier score
\[
\mathrm{BS}(p,y) := (y-p)^2.
\]
Compared to log-loss, the Brier score is less sensitive to extreme
predictions near $0$ and $1$, making it more stable when evaluating
local corrections.

We define the proxy metric as the corrected validation Brier score
\begin{equation}
\label{eq:proxy_bs}
\mathrm{Proxy}(\sigma,\phi)
:=
\frac{1}{n}\sum_{i=1}^n
\Big(
y_i - \tilde{f}_{\text{raw}}(x_i)
\Big)^2.
\end{equation}
Lower values of $\mathrm{Proxy}(\sigma,\phi)$ indicate that the learned
local correction improves predictive accuracy.

This quantity depends only on $(f_i, y_i)$ and
$\widehat{\delta}_\phi(x_i)$, and can therefore be computed without
access to $\delta(x)$ or $\eta(x)$.

The proxy admits a natural statistical interpretation. The Brier score
is a strictly proper scoring rule, so in expectation it is minimised by
the true conditional probability $\eta(x)$. Consequently, if
$\widehat{\delta}_\phi(x)$ is aligned with the true miscalibration
field, the corrected predictor $\tilde{f}(x)$ will improve validation
Brier score. Minimizing $\mathrm{Proxy}(\sigma,\phi)$ therefore favours
hyperparameters that yield locally informative and well-aligned estimates of the miscalibration field.

\section{Proof of Proposition~\ref{prop:variance}}
\label{app:variance}

We prove the variance bound for a fixed representation $\phi$. Recall the
estimator
\[
\widehat{\delta}_\phi(x)
=
\frac{\sum_{i=1}^n K(\phi(x),\phi(x_i))\,r_i}
     {\sum_{j=1}^n K(\phi(x),\phi(x_j))}.
\]
For a fixed query point $x$, define the normalised kernel weights
\[
w_i(x)
:=
\frac{K(\phi(x),\phi(x_i))}
{\sum_{j=1}^n K(\phi(x),\phi(x_j))}
=
\frac{K(\phi(x),\phi(x_i))}{m(x)},
\]
where
\[
m(x)=\sum_{j=1}^n K(\phi(x),\phi(x_j))
\]
is the effective neighbourhood mass. Conditional on the inputs
$\{x_i\}_{i=1}^n$ and the fixed representation $\phi$, the weights
$w_i(x)$ are deterministic, non-negative, and satisfy
$\sum_i w_i(x)=1$. Hence
\[
\widehat{\delta}_\phi(x)=\sum_{i=1}^n w_i(x)r_i.
\]

Since the samples are i.i.d., the labels are conditionally independent given
the inputs. Also, $f_i=f(x_i)$ is deterministic given $x_i$, so
\[
\Var(r_i\mid x_i)
=
\Var(y_i-f(x_i)\mid x_i)
=
\Var(y_i\mid x_i).
\]
Under the Bernoulli model
$Y\mid X=x\sim\mathrm{Bernoulli}(\eta(x))$,
\[
\Var(r_i\mid x_i)
=
\eta(x_i)(1-\eta(x_i))
\le \frac14.
\]
More generally, suppose $\Var(r_i\mid x_i)\le C_0$ for all $i$. Then
\[
\begin{aligned}
\Var\!\left(\widehat{\delta}_\phi(x)\mid \{x_i\}_{i=1}^n,\phi\right)
&=
\Var\!\left(\sum_{i=1}^n w_i(x)r_i
\,\middle|\, \{x_i\}_{i=1}^n,\phi\right)  \\
&=
\sum_{i=1}^n w_i(x)^2 \Var(r_i\mid x_i) \\
&\le
C_0\sum_{i=1}^n w_i(x)^2 .
\end{aligned}
\]
If the kernel is bounded as $0\le K(\cdot,\cdot)\le K_{\max}$, then
\[
\sum_{i=1}^n w_i(x)^2
=
\frac{\sum_{i=1}^n K(\phi(x),\phi(x_i))^2}{m(x)^2}
\le
\frac{K_{\max}\sum_{i=1}^n K(\phi(x),\phi(x_i))}{m(x)^2}
=
\frac{K_{\max}}{m(x)}.
\]
Therefore,
\[
\Var\!\left(\widehat{\delta}_\phi(x)\mid \{x_i\}_{i=1}^n,\phi\right)
\le
\frac{C_0K_{\max}}{m(x)}.
\]
Setting $C=C_0K_{\max}$ proves the claim.

\paragraph{Remark.}
The bound treats $\phi$ as fixed, isolating the variance-reduction effect of
kernel averaging in a chosen geometry. It shows that the estimator variance
decreases inversely with the effective neighbourhood mass $m(x)$. In contrast,
individual residuals have variance
$\Var(r_i\mid x_i)=\eta(x_i)(1-\eta(x_i))=\mathcal{O}(1)$, which does not
diminish with sample size in the absence of repeated observations at the same
input. In the full algorithm, $\phi$ is learned from residuals, so the
neighbourhood-mass regulariser in Eq.~\eqref{eq:loss} and validation-based
model selection are needed to avoid degenerate geometries with $m(x)\approx 1$
that fit pointwise Bernoulli noise.

This result is therefore not a recovery guarantee for the learned geometry.
We include it to clarify the statistical role of kernel averaging and
neighbourhood mass, rather than as a main theoretical contribution. The
question of when the learned representation recovers the true calibration
geometry remains an open theoretical problem.

\section{Implementation Details}
\label{app:implementation}

% \subsection{Reproducibility}
% To facilitate reproducibility, we provide an anonymised repository containing the code, experiment configurations, and instructions required to reproduce the reported results: \url{https://anonymous.4open.science/r/miscalibration-discovery-0974}.

\subsection{Representation learning network}
The network $\phi$ is a feedforward MLP with $L$ hidden layers
of width $h$, GELU activations, and an output projection of dimension
$d_{\text{out}}$ followed by $\ell_2$-normalisation. Its capacity differs between the
synthetic and real-data settings to reflect the input dimensionality:
\begin{itemize}[left=1em]
    \item \textbf{Synthetic experiments.} The input is a 2-dimensional
    embedding ($d_{\text{in}}=2$). We use $h=256$, $L=2$,
    $d_{\text{out}}=64$.
    \item \textbf{LLM experiments.} The input is the final-layer
    last-token hidden state of the frozen base model
    ($d_{\text{model}}$). We use $h=512$, $L=2$, $d_{\text{out}}=128$.
    $d_{\text{model}}$ varies by model family and size.
\end{itemize}
A dropout rate of $0.1$ is applied between hidden layers.

\subsection{Training objective}
We minimise the kernel-discovery loss
\begin{equation*}
\mathcal{L}(\phi)
=
-
\frac{1}{n}\sum_{i=1}^n \widehat{\delta}_\phi(x_i)^2
+
\lambda
\frac{1}{n}\sum_{i=1}^n
\left[\max(0,m_{\min}-m_i)\right]^2,
\end{equation*}
where 
\begin{equation*}
\widehat{\delta}_\phi(x)
=
\frac{\sum_{i=1}^n K_\sigma(\phi(x),\phi(x_i))\, r_i}
     {\sum_{i=1}^n K_\sigma(\phi(x),\phi(x_i))},  \quad 
m_i
:=
\sum_{j=1}^n K_\sigma(\phi(x_i),\phi(x_j))
\end{equation*}
using the Gaussian kernel:
\begin{equation*}
K_\sigma(\phi(x),\phi(x'))
=
\exp\!\left(
-\frac{\|\phi(x)-\phi(x')\|^2}{\sigma^2}
\right).    
\end{equation*}
We set $m_\text{min}=20$ and optimise the penalty strength $\lambda$ and the kernel bandwidth $\sigma$ via hyperparameter tuning.
Checkpoint selection and hyperparameter tuning use the corrected validation Brier score computed on the validation split, as described in~\ref{app:exp:proxy_selection}.

\subsection{Optimisation}
All models are trained with Adam ($\text{lr}=3\!\times\!10^{-5}$,
weight decay $7\!\times\!10^{-6}$), batch size $1024$, gradient-norm
clipping at $1.0$, and a maximum of $100$ epochs. We use early stopping 
with patience $20$, and restore the best checkpoint by the validation
metric. Training runs on a single A100 NVIDIA GPU.

\subsection{Computational Cost and Scalability}
\label{app:computational_cost}

The main computational cost of the method comes from evaluating kernel
neighbourhoods. Given a query set of size $n_{\mathrm{eval}}$ and a training
neighbour bank of size $n_{\mathrm{bank}}$, exact smoothing against the bank
has time complexity
$O(n_{\mathrm{eval}} n_{\mathrm{bank}} d_{\mathrm{out}})$,
where $d_{\mathrm{out}}$ is the dimension of the learned representation
$\phi(x)$. In our experiments, the neighbour bank is a uniformly sampled subset
of the training split capped at $n_{\mathrm{bank}}=20{,}000$ examples. Thus,
the evaluation cost scales linearly with the evaluation set size and with the
capped bank size.

In our implementation, smoothing is exact with respect to the sampled bank but
chunked over query points. For a query chunk of size $Q$, we compute the
$Q\times n_{\mathrm{bank}}$ distance matrix, apply the Gaussian kernel, and
multiply by the bank residuals. Thus, the peak memory cost is
$O(Qn_{\mathrm{bank}})$ rather than
$O(n_{\mathrm{eval}}n_{\mathrm{bank}})$, and the full query-by-bank kernel
matrix is never materialized.

During representation learning, the quadratic kernel computation is performed
within minibatches rather than over the full training set. For minibatch size
$B$, computing all pairwise representation-space kernels costs
$O(B^2 d_{\mathrm{out}})$ time and $O(B^2)$ memory. The quadratic cost
is controlled by the minibatch size, while validation and test evaluation use
the capped training neighbour bank.

All experiments in this paper use exact kernel evaluations with respect to a
uniformly sampled training bank capped at $20{,}000$ examples. We choose this
simple bank-sampling strategy because it preserves the training distribution in
expectation and fixes the maximum cost of evaluation at
$O(n_{\mathrm{eval}}\cdot 20{,}000 \cdot d_{\mathrm{out}})$. The resulting
estimator can be viewed as a Monte Carlo approximation to full-bank smoothing.
Alternative scaling strategies include restricting the kernel average to the
top-$k$ nearest neighbours in the learned representation~\citep{stone1977consistent,indyk1998approximate}.

As a representative synchronized runtime on a single A100 GPU, for a setting
with $n_{\mathrm{test}}=9{,}103$, $n_{\mathrm{train}}\approx70\mathrm{k}$,
$n_{\mathrm{bank}}=20{,}000$, and $d_{\mathrm{out}}=128$, training the
representation map $\phi$ took approximately $40$ minutes. Computing
$\phi(x)$ for the test split took $4.7$ seconds, building the sampled
neighbour bank took $10.2$ seconds, and computing test-set field values by
exact kernel averaging against the sampled bank took $0.01$ seconds.

\subsection{Datasets}

\paragraph{Synthetic data.} Construction of the synthetic datasets is described in Appendix~\ref{app:exp:data}.

\paragraph{Real data.}
For the real-data experiments we use four real-world datasets: HH-RLHF, MedMCQA, MMLU, MMLU-Pro, each scored under one of twelve base model checkpoints from Hugging Face: Qwen3-1.7B-Base, Qwen3-4B-Base, Qwen3-8B-Base, Ministral-3-3B-Base-2512, Ministral-3-8B-Base-2512, Ministral-3-14B-Base-2512, Llama-3.2-1B, Llama-3.2-3B, Llama-3.1-8B,  gemma-3-1b-pt, gemma-3-4b-pt, gemma-3-12b-pt.
For each model--dataset pair we precompute (i) the last-token hidden state of the base model on
the prompt, and (ii) the model's predicted confidence $f(x)$.

For HH-RLHF, confidence values indicate the model-assigned probability that the first of two candidate responses is preferred.
The label $y$ is therefore $y=1$ if the ground-truth human preference label indicates the first response and
$y=0$ otherwise. For the remaining multiple-choice datasets, we define the
binary variable as the correctness of the model's selected answer:
$y=1$ if the answer with highest model probability is correct and $y=0$
otherwise. The associated confidence score $f(x)$ is the model probability
assigned to this selected answer. Sample sizes range from $\approx 12\text{k}$
(MMLU-Pro) to $\approx 193\text{k}$ (MedMCQA). All datasets are split
$80/10/10$ into train/validation/test by a random permutation.

\paragraph{Prompt formats for eliciting LLM confidence scores.}
To elicit the LLM confidence scores we use the prompt formats presented in
Prompt~\ref{prompt:mcqa} and \ref{prompt:hhrlhf}. We treat LLMs as black-box probabilistic classifiers whose output is a
distribution over a fixed set of candidate completions
$\mathcal{V}$ (the answer letters). For each candidate
$v \in \mathcal{V}$ we compute the conditional log-likelihood
$\log p_{\text{LLM}}(v \mid \text{prompt})$ as the sum of the token-level
log-probabilities of $v$ and normalize across candidates with a softmax:
\[
    \tilde{p}(v \mid \text{prompt})
    \;=\;
    \frac{\exp\bigl(\log p_{\text{LLM}}(v \mid \text{prompt})\bigr)}
         {\sum_{v' \in \mathcal{V}} \exp\bigl(\log p_{\text{LLM}}(v' \mid \text{prompt})\bigr)}.
\]
This restriction to $\mathcal{V}$ removes probability mass that the LLM
would otherwise assign to off-task tokens (whitespace, alternative
phrasings of the answer, etc.) and yields a well-defined Bernoulli /
categorical predictive distribution that we can study.

\textbf{MMLU, MMLU-Pro, MedMCQA.} For these multi-choice datasets we use
Prompt~\ref{prompt:mcqa}. The candidate set is the set of answer letters
$\mathcal{V} = \{\text{A}, \text{B}, \text{C}, \dots\}$
($|\mathcal{V}| = 4$ for MedMCQA, between $4$ and $10$ for MMLU and
MMLU-Pro). The base predictor is the LLM's confidence in its top choice,
\[
    f(x) \;=\; \max_{v \in \mathcal{V}} \tilde{p}(v \mid \text{prompt}(x)),
\]
and the binary outcome is the correctness indicator
\[
    y \;=\; \mathbbm{1}\bigl[\arg\max_{v \in \mathcal{V}}
                  \tilde{p}(v \mid \text{prompt}(x))
                  \;=\; v^\star(x)\bigr],
\]
where $v^\star(x)$ is the gold answer letter. Calibration is therefore
assessed on the LLM's \emph{top-1} confidence.

\textbf{HH-RLHF (preference judging).} We re-purpose the
Anthropic HH-RLHF dataset as a binary
preference-judging task. For each example we extract the user prompt and
the chosen / rejected response pair, then assign them to slots
A and B uniformly at random; the binary label
$y = \mathbbm{1}[\text{option A is the chosen response}]$. We then prompt
the LLM with Prompt~\ref{prompt:hhrlhf} so that the next token must be
either ``A'' or ``B''. The candidate set is
$\mathcal{V} = \{\text{A}, \text{B}\}$ and we take
\[
    f(x) \;=\; \tilde{p}(\text{A} \mid \text{prompt}(x))
\]
as the base predictor, paired directly with $y$. A well-calibrated judge should satisfy
$\mathbb{P}[y = 1 \mid f(x) = f] = f$ across the full
$[0, 1]$ range.

Since candidate completions are single tokens, in both cases the sequence of token log-probabilities is computed via a
single forward pass per candidate completion (no decoding). Prompt embeddings $x$ used for the calibration network $\phi$ are
extracted from this same forward pass as the last hidden state at the last input token.

\begin{prompt}[h]
\begin{promptbox}{}
Question: {question}

{choices}

The correct answer is:
\end{promptbox}
\vspace{-1em}
\caption{Prompt for eliciting LLM confidence scores on the multi-choice datasets (MMLU, MMLU-Pro, MedMCQA)}
\label{prompt:mcqa}
\end{prompt}

\begin{prompt}[h]
\begin{promptbox}{}
Act as an impartial judge and evaluate the quality of two candidate responses (Response A and Response B) to the user query displayed below. Make your decision based on which response is more helpful while not being harmful.
    
## Context and Query:
{prompt}

## Response A:
{response_a}

## Response B:
{response_b}

## Your Choice:
In this context, the Assistant should respond with Response 
\end{promptbox}
\vspace{-1em}
\caption{Prompt for eliciting LLM confidence scores on the HH-RLHF dataset.}
\label{prompt:hhrlhf}
\end{prompt}

\subsection{Diagnositc controls for Recalibration}
We compare $\hat{\delta}_\phi$-based correction to two standard
post-hoc calibrators:
\begin{itemize}[left=1em]
    \item \textbf{Temperature scaling.} A single scalar $T$ is fit by
    full-batch L-BFGS optimisation with respect to the binary negative log likelihood.
    \item \textbf{Isotonic regression.} We use the off-the-shelf implementation from
    \texttt{scikit-learn}.
\end{itemize}
Both methods are fit on the training splits of datasets.

\subsection{Licenses and terms of use for existing assets.}
All datasets and model checkpoints used in this work are existing public
assets, and we cite the corresponding original papers in the main text and
references. We use HH-RLHF, MedMCQA, MMLU, and MMLU-Pro under their respective
public dataset terms, and we use publicly available Hugging Face checkpoints
for Qwen3, Ministral-3, Llama-3.1/3.2, and Gemma-3 under the licenses and
acceptable-use terms specified on their model cards. We do not redistribute the
raw datasets or model weights; the released code contains experiment
configurations and scripts for reproducing the analyses from the original
sources. Users of the released code are responsible for complying with the
licenses and terms of the underlying datasets and checkpoints.

\section{Experiments: Synthetic Illustrations (Extended Results)}\label{app:exp}

\subsection{Datasets and Experimental Setup}\label{app:exp:data}

We construct two low-dimensional datasets in which inputs from distinct
regions of the input space exhibit different levels of miscalibration.

\paragraph{3-clusters.}  Inputs $X \in \mathbb{R}^2$ are sampled from a mixture of three Gaussian clusters. Predicted confidence scores are generated from a latent logit function  $\ell(x) =w^\top x  + \epsilon_\ell$, $\epsilon_\ell \sim \mathcal{N}(0, \sigma_\ell^2)$ where $w \in \mathbb{R}^2$ is a fixed direction. The observed confidence scores are defined as $f(x) = \text{sigm}(\ell(x))$, where $\text{sigm}(\cdot)$ denotes the logistic sigmoid. This construction ensures that
prediction confidences are input-dependent while remaining smoothly distributed in $[0,1]$ across clusters.
To introduce systematic miscalibration, we assign each cluster $c$ a logit shift $\Delta_c \in \{-1, 0, +1\}$, representing the strength and
direction of the calibration bias. The true conditional probability is then defined as $\eta(x) = \text{sigm}\!\big(\ell(x) + \Delta_{c(x)}\big)$ where $c(x)$ denotes the cluster membership of $x$. Finally, labels are sampled as $Y \mid X=x \sim \mathrm{Bernoulli}(\eta(x))$. The observed dataset therefore consists of triples $(x_i, f_i, y_i)$. This construction induces a ground-truth miscalibration field $\delta(x) = \eta(x) - f(x)$, as visualised in the left pane of Figure~\ref{fig:cluster_scatter}. The field is piecewise-structured in the input space but invisible to score-based calibration diagnostics that implicitly aggregate across the entire input space.

\paragraph{Sinusoidal field.} To study recovery of smoothly varying miscalibration, we construct a
dataset where the miscalibration field varies continuously over the
input space. Inputs are sampled uniformly from the unit square. Predicted confidence scores are generated from a latent
logit model as in the 3-clusters dataset. We introduce input-dependent miscalibration through a sinusoidal logit
shift field defined directly over the input space: $\Delta(x) = A \sin\!\big(2\pi k \, u^\top x\big)$,
where $u \in \mathbb{R}^2$ is a unit direction vector, $k \in \mathbb{N}$
controls the spatial frequency of regime alternation, and $A > 0$
controls the strength of miscalibration. The true conditional
probability is then given by $\eta(x) = \text{sigm}\!\big(\ell(x) + \Delta(x)\big)$, 
and labels are sampled as $Y \mid X=x \sim \mathrm{Bernoulli}(\eta(x))$. 
The observed dataset consists of triples $(x_i, f_i, y_i)$.
This construction induces a smooth ground-truth miscalibration field $\delta(x) = \eta(x) - f(x)$, 
which oscillates across the input space with frequency $k$ (see Figure~\ref{fig:sinusoid_scatter}).
The sinusoidal field provides a controlled setting to evaluate
whether the learned representation $\phi$ can recover smoothly varying,
input-dependent miscalibration from noisy observations.

\paragraph{Setup.} For each dataset, we generate $n=10{,}000$ samples and split them into
train/validation/test sets with proportions $80/10/10$. The
representation $\phi$ is implemented as a two-layer MLP trained using
the objective in Eq.~(\ref{eq:loss}). At validation and test time, the local miscalibration field
$\widehat{\delta}_\phi(x)$ is estimated via kernel smoothing in the
learned representation space (Eq.~(\ref{eq:DeltaHat})) using the training
set as the neighbour bank (i.e., kernel averages are computed over
training points only). To quantitatively assess recovery of the underlying structure, we
measure the Pearson correlation between the estimated field
$\widehat{\delta}_\phi(x)$ and the ground-truth field $\delta(x)$ on the
test set.

\subsection{Results: Clustered Miscalibration Regimes}\label{app:exp:3clust_results}

\begin{figure}[h]
    \centering
    \subfloat[Left: True cluster-dependent miscalibration field $\delta(x)$. Right: Estimated miscalibration field $\widehat{\delta}_\phi(x)$ learned from $(x, f, y)$ only.]{
        \includegraphics[height=2.9cm]{figures/kernel_synthetic_3_clusters.pdf}
        \label{fig:cluster_scatter}
    }
    \hfill
    \subfloat[Global reliability diagram (first pane) suggests good calibration due to cancellation across regimes. Cluster-level analysis reveals systematic over- and underconfidence.]{
        \includegraphics[height=2.9cm]{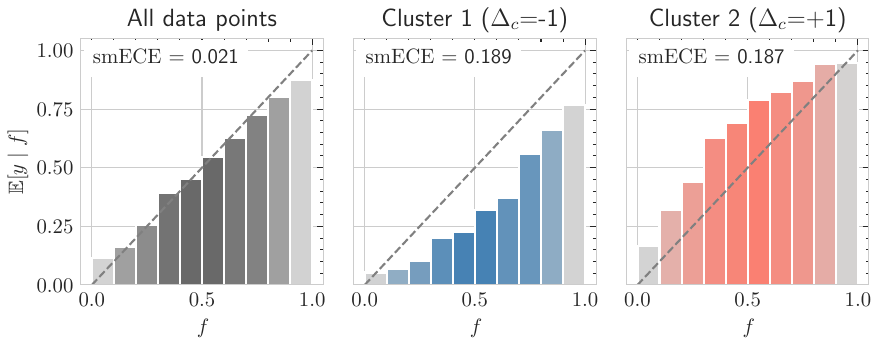}
        \label{fig:cluster_relplot}
    }
    \caption{Global calibration appears accurate due to cancellation. The learned field reveals cluster-dependent over- and underconfidence.}
    \label{fig:3clusters}
\end{figure}

Although each cluster exhibits a distinct and systematic calibration bias, the
global reliability diagram (Fig.~\ref{fig:cluster_relplot}, 1st pane) suggests
near-calibrated behaviour due to cancellation between overconfident and
underconfident regions. Cluster-wise reliability analysis
(Fig.~\ref{fig:cluster_relplot}, 2nd and 3rd panes) confirms, however, that systematic miscalibration is present. This highlights the limitation of a standard calibration analysis if appropriate slicing of the data is not known a priori.

In contrast, the learned estimator $\widehat{\delta}_\phi(x)$ closely
matches the spatial structure of the ground-truth miscalibration field.
As shown in Fig.~\ref{fig:cluster_scatter}, the method successfully
separates regions of overconfidence and underconfidence in the input
space despite having access only to $(x,f,y)$ and not to $\eta(x)$ or
cluster labels. Quantitatively, the estimated field $\widehat{\delta}_\phi(x)$ achieves a Pearson
correlation of $r=0.95$ with the ground-truth $\delta(x)$ on the test
set, demonstrating accurate recovery of both the sign and magnitude of
local miscalibration.

\subsection{Results: Smooth Miscalibration Field}\label{app:exp:sinusoid_results}

\begin{figure}[h]
    \centering
    \subfloat[Left: True miscalibration field $\delta(x)$. Right: Estimated miscalibration field $\widehat{\delta}_\phi(x)$. Sinusoidal dataset with $A=0.6$ and $k=3$.]{
        \includegraphics[height=3cm]{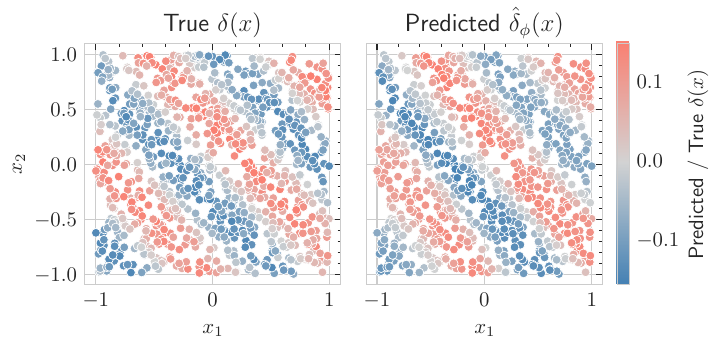}
        \label{fig:sinusoid_scatter}
    }
    \qquad
    \subfloat[Comparison of model performance across varying signal strengths $A$ and regime oscillation frequencies $k$.]{
        \quad
        \includegraphics[height=3cm]{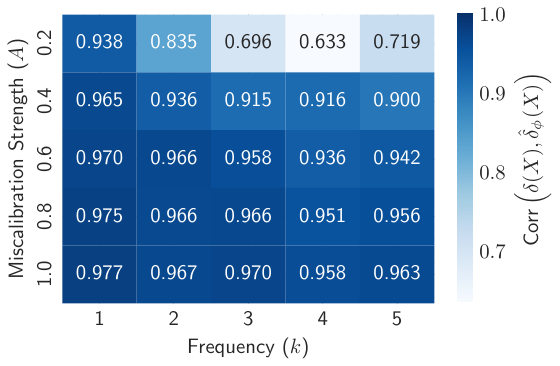}
        \quad
        \label{fig:sinusoidal_pcorr}
    }
    \caption{Recovering hidden calibration regimes for a continuous miscalibration field.}
    \label{fig:sinusoid}
\end{figure}

Figure~\ref{fig:sinusoid_scatter} shows recovery of a smoothly varying
miscalibration field for a representative setting $(A=0.6, k=3)$. The
estimated field $\widehat{\delta}_\phi(x)$ closely matches the spatial
oscillation pattern of the ground-truth $\delta(x)$, correctly capturing
both the sign and structure of overconfident and underconfident regions.

Figure~\ref{fig:sinusoidal_pcorr} reports the Pearson correlation between
$\widehat{\delta}_\phi(x)$ and the ground-truth field $\delta(x)$ across
different amplitudes $A$ and spatial frequencies $k$. Overall, the method
achieves consistently high correlation across a wide range of
difficulties. Performance degrades gradually as $k$ increases and $A$
decreases, reflecting the increased difficulty of recovering
higher-frequency and lower signal-to-noise miscalibration structure from
noisy Bernoulli residuals. Nevertheless, the estimator maintains strong
positive correlation even in challenging regimes, indicating robust
recovery of smooth input-dependent calibration patterns.

Overall, results from sections \ref{app:exp:3clust_results} and \ref{app:exp:sinusoid_results}
provide a visual and quantitative validation of the proposed approach: learning a calibration-aware representation
induces a geometry in which neighbourhood averaging yields stable and informative estimates of the input-dependent miscalibration field, uncovering hidden calibration regimes that are invisible to score-based diagnostics.

\subsection{Interaction Between Kernel Bandwidth and Mass Regularisation}
\label{app:exp:interaction}

\begin{figure}[h]
    \centering
    \includegraphics[height=3cm]{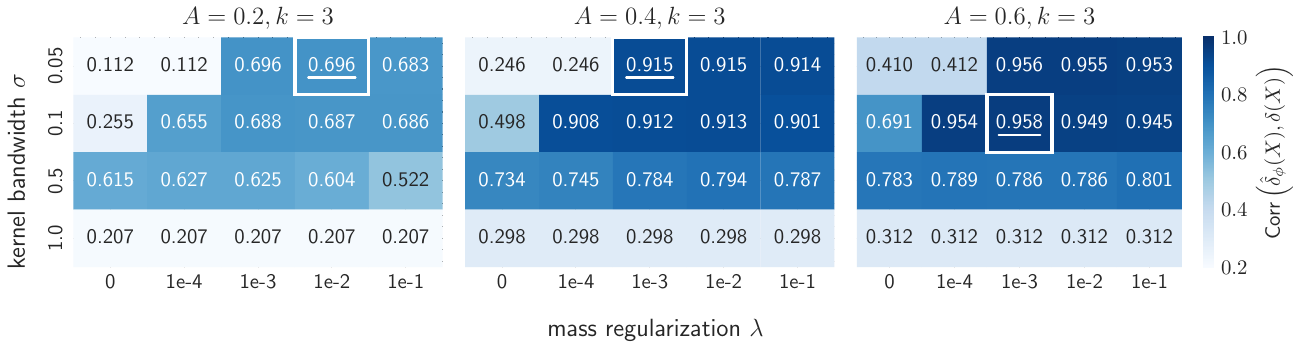}
    \caption{Interaction between kernel bandwidth $\sigma$ and mass regularisation $\lambda$ on sinusoidal miscalibration fields. Test Corr$(\widehat{\delta}_\phi(X), \delta(X))$ across a grid of $(\sigma,\lambda)$, with best settings per dataset highlighted. Small bandwidths offer higher spatial resolution but become unstable when $\lambda$ is too small due to low neighbourhood mass. Moderate regularisation stabilises performance, while large $\sigma$ is bias-dominated and largely insensitive to $\lambda$, highlighting the joint role of $(\sigma,\lambda)$.}
    \label{fig:sigma_lambda}
\end{figure}

Having established that the method can recover both discrete and smooth fields, we next study the sensitivity of recovery to the kernel bandwidth and mass regularisation parameters

Figure~\ref{fig:sigma_lambda} reports the Pearson correlation between
$\widehat{\delta}_\phi(X)$ and $\delta(X)$ on the test set across a grid
of kernel bandwidths $\sigma$ and mass regularisation strengths $\lambda$
for sinusoidal datasets with fixed frequency $k=3$ and varying amplitudes $A$.
Results reported for model checkpoints at the lowest validation loss during training.
We observe a pronounced interaction between the two hyperparameters.
Small bandwidths ($\sigma \in \{0.05, 0.1\}$) provide the spatial
resolution required to recover the oscillatory miscalibration field, but
become unstable when mass regularisation is too weak
($\lambda \leq 10^{-4}$), leading to performance degradation.
This behaviour is consistent with neighbourhood collapse and
high-variance residual estimates under low effective neighbourhood mass.
In contrast, moderate regularisation ($\lambda \in [10^{-3}, 10^{-2}]$)
stabilises the estimator and consistently yields the best performance in
low- and moderate-signal regimes.

\subsection{Proxy-Based Hyperparameter Selection Without Access to \texorpdfstring{$\delta(x)$}{deltax}}
\label{app:exp:proxy_selection}

The results in Fig.~\ref{fig:sigma_lambda} highlight that the choice of
kernel bandwidth $\sigma$ and mass regularisation $\lambda$ has a
significant impact on performance. In practice, however, the ground-truth
miscalibration field $\delta(x)$ is unknown, and oracle metrics such as
$\mathrm{Corr}(\widehat{\delta}_\phi(X), \delta(X))$ cannot be used for
model selection. Moreover, the training objective in
Eq.~(\ref{eq:loss}) depends explicitly on the kernel bandwidth $\sigma$
and the representation $\phi$, and is therefore not directly comparable
across different hyperparameter settings. This raises a central practical question: can
hyperparameters be reliably selected using a proxy criterion computed
only from the observed triples $(x,f,y)$? To address this challenge we use a validation-based proxy metric defined
as the Brier score of the locally corrected predictions
$f(x) + \widehat{\delta}_\phi(x)$; a formal definition and discussion
are provided in Appendix~\ref{app:method:proxy_metric}.

To investigate this, we leverage the synthetic sinusoidal setting where
the oracle performance $O(\sigma,\lambda) := \mathrm{Corr}\!\big(\widehat{\delta}_\phi(X), \delta(X)\big)$
can be computed on the test set for evaluation purposes only. For each
dataset instance (defined by amplitude $A$ and frequency $k$), we
evaluate a grid of hyperparameters $(\sigma,\lambda)$ and compare: (i)
the proxy score used for model selection (computed on validation data),
and (ii) the oracle performance $O(\sigma,\lambda)$ (computed on the test set).

To quantify the quality of proxy-based selection we define three diagnostic
quantities. We compute the \emph{oracle spread}
$\Delta O := O_{\max} - O_{\min}$ over the hyperparameter grid, which
measures how distinguishable different configurations are. Second, we
measure the \textit{Spearman rank correlation} between proxy scores and oracle
performance across the grid. Third, we report the \emph{selection regret} $\mathrm{Regret}
:=
O^* - O(\hat{\sigma},\hat{\lambda}),
$
where $O^* = \max_{\sigma,\lambda} O(\sigma,\lambda)$ and
$(\hat{\sigma},\hat{\lambda})$ denotes the hyperparameters corresponding to the best proxy value on the validation set.

\begin{figure}[h]
    \centering
    \begin{minipage}[t]{0.5\linewidth}
        \vspace{0pt}
        \centering
        \includegraphics[width=\linewidth]{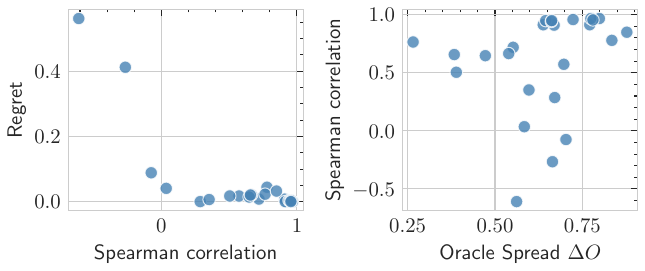}
        \caption{Proxy ranking diagnostics. Left: Spearman correlation vs.\ regret.
        Right: oracle spread vs.\ Spearman correlation.}
        \label{fig:proxy_selection_scatter_app}
    \end{minipage}
    \hfill
    \begin{minipage}[t]{0.4\linewidth}
        \vspace{0pt}
        \centering
        \captionof{table}{Oracle regret $O^* - O(\hat{\sigma},\hat{\lambda})$
        across amplitudes $A$ and frequencies $k$ (lower is better).}
        \label{tab:proxy_regret}
        \small
        \setlength{\tabcolsep}{3pt}
        \begin{tabular}{c|ccccc}
            \toprule
            $A \backslash k$ & 1 & 2 & 3 & 4 & 5 \\
            \midrule
            0.2 & 0.088 & 0.412 & 0.041 & 0.562 & 0.000 \\
            0.4 & 0.006 & 0.002 & 0.007 & 0.017 & 0.000 \\
            0.6 & 0.008 & 0.020 & 0.000 & 0.044 & 0.000 \\
            0.8 & 0.014 & 0.023 & 0.003 & 0.032 & 0.000 \\
            1.0 & 0.017 & 0.020 & 0.002 & 0.003 & 0.000 \\
            \bottomrule
        \end{tabular}
    \end{minipage}
\end{figure}

Figure~\ref{fig:proxy_selection_scatter_app} (left) shows a positive
relationship between Spearman correlation and selection regret. Higher
rank agreement (lower value of the proxy is better) between proxy and oracle scores leads to lower regret,
indicating that improved proxy ranking translates directly into
near-optimal hyperparameter selection. Figure~\ref{fig:proxy_selection_scatter_app} (right) shows a clear 
relationship between oracle spread and Spearman rank correlation. When
the oracle surface has a large spread, the proxy reliably ranks
hyperparameter configurations; when the spread is small, the oracle
surface is flatter and rank agreement deteriorates due to
intrinsic ambiguity in the selection problem.

Finally, Table~\ref{tab:proxy_regret} reports regret across amplitudes
$A$ and frequencies $k$. Overall, the median regret across all settings
remains low ($\approx0.008$), and is near zero for moderate-to-high signal strengths
($A \ge 0.4$), indicating that proxy-based tuning frequently recovers
near-oracle hyperparameters. Larger regret occurs primarily in
low-signal settings ($A=0.2$) and higher frequencies, where the oracle
surface is flatter and the estimation problem becomes more
noise-dominated.

Overall, these results suggest that proxy-based selection is reliable
whenever the miscalibration structure is identifiable. In low-signal
regimes, reduced rank agreement coincides with small oracle spread,
implying that hyperparameter selection is inherently less stable rather
than systematically misled by the proxy.

\subsection{An Alternative Approach: Residual Regression}
\label{app:exp:resreg}

A natural alternative to learning a calibration geometry is to directly
estimate the miscalibration field $\delta(x)$ by regressing the empirical
residuals $r_i$ on the input $x$. Since $\delta(x) = \mathbb{E}[R \mid X=x]$,
one can train a predictor $g_\theta$ by minimizing
\begin{equation}\label{eq:resreg}
    \min_{\theta}\ \frac{1}{n}\sum_{i=1}^n \big(r_i - g_\theta(x_i)\big)^2.
\end{equation}
We refer to this method as \emph{residual regression} (ResReg).
In contrast to our approach, ResReg attempts to learn a pointwise
predictor of a high-variance target, whereas the proposed kernel method
reduces variance through neighbourhood averaging in a learned geometry
(Proposition~\ref{prop:variance}).

In this section, we compare the proposed kernel-based estimator with the
ResReg baseline on the sinusoidal dataset introduced earlier. The ResReg model is implemented as an MLP that
takes $x$ as input and directly predicts the residual
$r(x)=y-f(x)$, trained with mean squared error as in Eq.~\ref{eq:resreg}. At test time, we evaluate
the Pearson correlation between model predictions and the
ground-truth miscalibration field $\delta(x)$.

\begin{figure}[h]
    \centering
    \subfloat[Test set $\text{Corr}(\left(g_\theta(X), \delta(X)\right)$ vs. complexity for varying $k$ (fixed $A=1.0$). Increasing regime frequency substantially raises the capacity required for ResReg, with all models performing near chance for large $k$.]{
        \quad
        \includegraphics[height=3cm]{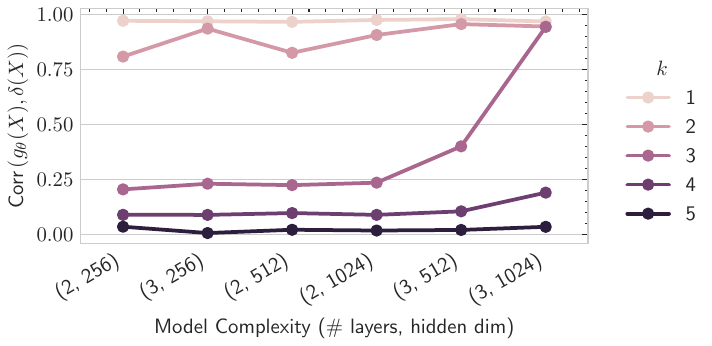}
        \label{fig:resreg_capacity}
        \quad
    }
    \qquad
    \subfloat[ResReg model performance across varying signal strengths $A$ and regime oscillation frequencies $k$. ]{
        \quad
        \includegraphics[height=3cm]{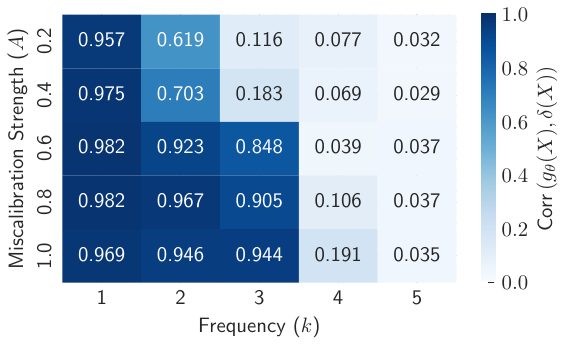}
        \quad
        \label{fig:resreg_axk}
    }
    \caption{Pointwise residual regression is more sensitive to regime complexity and noise.}
    \label{fig:resreg_vs_kernel}
\end{figure}

\paragraph{ResReg requires increasing capacity as $k$ increases.}
Figure~\ref{fig:resreg_capacity} reports performance for amplitude
$A=1.0$ across increasing regime frequencies $k$ and model capacities.
While residual regression performs well for low-frequency fields
($k=1,2$), performance deteriorates rapidly as the oscillation frequency
increases. Even large networks struggle to recover the structure for
$k\geq 3$, and all models fail for $k=4,5$, where correlations collapse
towards zero. We additionally observe that ResReg requires a substantially larger
training budget to achieve meaningful improvements. In our experiments,
ResReg models were trained for 300 epochs, whereas the kernel models
typically converged within 100 epochs. This indicates that directly learning a high-frequency
miscalibration field from noisy Bernoulli residuals is a task requiring high model complexity.

\paragraph{ResReg struggles in the low-signal, high-frequency regimes.}
Figure~\ref{fig:resreg_axk} shows the performance of the highest-capacity
model (3 layers, width 1024, 300 training epochs) across amplitudes $A$ and frequencies $k$.
Performance improves with increasing $A$, reflecting the higher
signal-to-noise ratio in the residuals. However, even at strong signal
levels ($A=1.0$), the model fails to recover high-frequency regimes
($k\geq 4$), with correlations remaining close to zero. In contrast, the kernel model (compare Figure~\ref{fig:resreg_axk} and Figure~\ref{fig:sinusoidal_pcorr}) maintains strong performance across all regimes despite a less complex architecture (2 layers, width 256) and lower training budget (100 epochs), suggesting greater robustness to low signal-to-noise conditions and miscalibration complexity. 

 Overall, the results reveal a fundamental difference between the two
approaches. Kernel smoothing aggregates residuals over local
neighbourhoods, yielding stable estimates even for high-frequency and
noisy miscalibration fields. In contrast, residual regression must learn
a pointwise mapping from high-variance targets, making it increasingly
sensitive to regime complexity and signal-to-noise ratio, consistent
with the variance reduction analysis in Proposition~\ref{prop:variance}.

\section{Additional Results for the Large Scale LLM Study}\label{app:exp_real}

\subsection{Characterisation of the Learned Miscalibration Fields}\label{app:exp_real:characterisation}

\begin{figure}[h]
    \centering
    \includegraphics[width=0.8\linewidth]{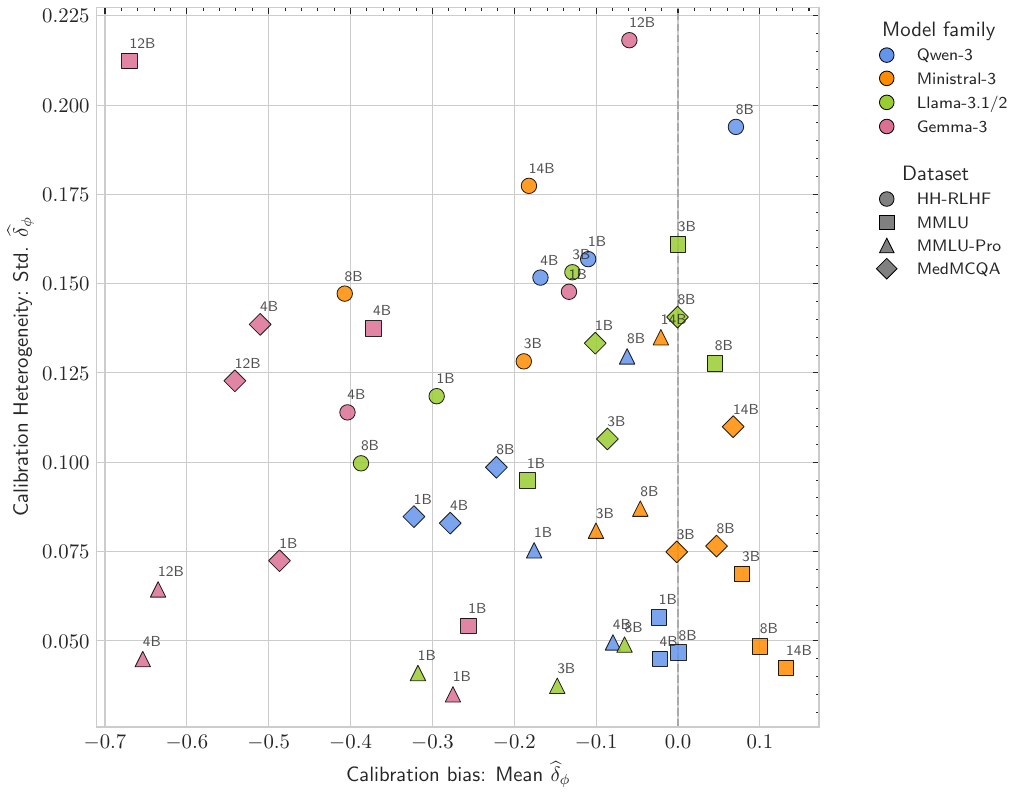}
    \caption{\textit{Miscalibration map.} Mean and standard deviation of the learned miscalibration fields $\widehat{\delta}_\phi$. Annotated with model size.}
    \label{fig:placeholder}
\end{figure}

\subsection{Raw vs. Learned Representation Space}\label{app:exp_real:raw_vs_learned}

One possible explanation for the discovered calibration regimes is that they are already present in the frozen LLM representation, and that learning a calibration-aware map $\phi$ is unnecessary. To test this, we compare our learned geometry against a frozen-embedding smoothing baseline. This baseline replaces the learned representation $\phi(x)$ with the raw final-layer hidden state $x$ of the frozen model, while keeping the rest of the estimation pipeline unchanged. In particular, we use the same residuals $r_i = y_i - f_i$, the same kernel estimator, the same training-set neighbour bank for validation and test points, and the same proxy validation-based bandwidth selection procedure.

Formally, for the frozen baseline we estimate
\begin{equation}
\hat{\delta}_{\mathrm{raw}}(x)
=
\frac{
\sum_{i \in \mathcal{D}_{\mathrm{train}}}
K_{\sigma_\text{raw}}(x, x_i) r_i
}{
\sum_{i \in \mathcal{D}_{\mathrm{train}}}
K_{\sigma_\text{raw}}(x, x_i)
},
\end{equation}
whereas our method estimates
\begin{equation}
\hat{\delta}_{\phi}(x)
=
\frac{
\sum_{i \in \mathcal{D}_{\mathrm{train}}}
K_{\sigma}(\phi(x), \phi(x_i)) r_i
}{
\sum_{i \in \mathcal{D}_{\mathrm{train}}}
K_{\sigma}(\phi(x), \phi(x_i))
}.
\end{equation}
The only difference between the two estimators is whether neighbourhoods are defined in the frozen LLM hidden-state geometry or in the learned calibration geometry.

\begin{figure}[h]
    \vspace{-1em}
    \centering
    \hspace{-1em}
    \subfloat[\textit{Frozen vs. learned calibration geometry.} Each point corresponds to one model--dataset pair. Kernel smoothing in the raw hidden-state space yields low-variance miscalibration fields that are mostly negative, capturing broad overconfidence but little input-dependent heterogeneity. In contrast, the learned calibration geometry produces higher-variance signed fields, revealing local variations in calibration behaviour.]{
        \hspace{0.75em}
        \includegraphics[height=2.8cm]{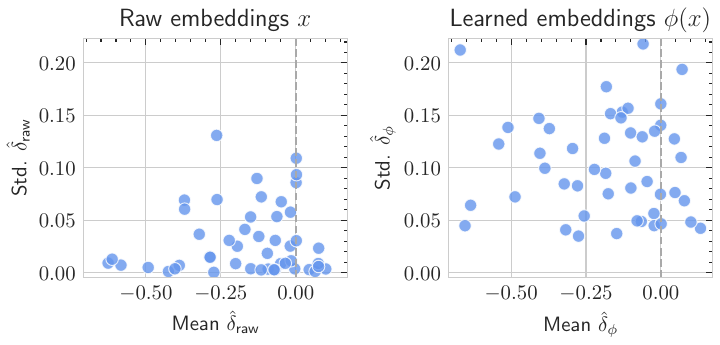}
        \label{fig:raw-vs-learned-fields}
        \hspace{0.75em}
    }
    \quad
    \subfloat[\textit{Worst-slice gaps under frozen and learned geometries.}  The first two panels show the distribution of
    $\mathrm{gap} = \mathrm{smECE}_{\mathrm{worst}}-\mathrm{smECE}_{\mathrm{global}}$
    for slices discovered using raw embeddings and learned embeddings, respectively.
    The learned geometry more frequently identifies held-out slices whose calibration error exceeds the global smECE. The right panel shows the paired difference $\mathrm{gap}_\phi - \mathrm{gap}_{\mathrm{raw}}$ across matched model--dataset pairs. ]{
        \includegraphics[height=2.8cm]{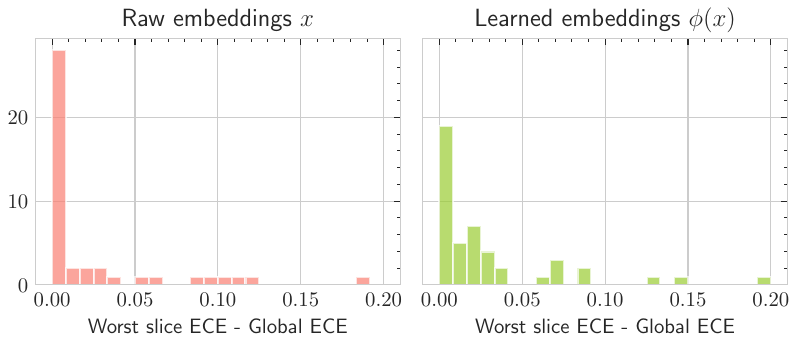}
        \label{fig:raw-vs-learned-gap}
    }
    \caption{}
    \label{fig:raw_vs_learned}
\end{figure}

Figure~\ref{fig:raw-vs-learned-fields} compares the distributional structure of the learned fields across model--dataset pairs. Smoothing in the raw embedding space produces fields with relatively low variance and predominantly negative mean values. This indicates that the frozen LLM geometry primarily captures a broad overconfidence bias: nearby examples in the raw representation tend to average to similar, mostly negative residuals. In contrast, the learned calibration geometry produces substantially higher variation in $\hat{\delta}_{\phi}$ across inputs. The learned fields also contain both strongly negative and positive regions, indicating that the representation has reorganised examples according to signed calibration behaviour rather than simply preserving the model's original semantic geometry.

We next ask whether this additional variation corresponds to meaningful calibration structure. For each geometry, we define overconfident and underconfident regions using the sign and magnitude of the corresponding estimated field, and compute the worst-slice calibration gap
\begin{equation}
\mathrm{gap}
=
\mathrm{smECE}_{\mathrm{worst}}
-
\mathrm{smECE}_{\mathrm{global}},
\end{equation}
where $\mathrm{smECE}_{\mathrm{worst}}$ is the larger smECE among the discovered overconfident and underconfident slices. A positive value of $\mathrm{gap}$ indicates that the geometry has identified a region whose calibration error is larger than suggested by the aggregate reliability metric.

Figure~\ref{fig:raw-vs-learned-gap} shows that raw embeddings often fail to reveal such hidden gaps: many model--dataset pairs have $\mathrm{gap} \approx 0$ under the frozen geometry. In contrast, the learned geometry more frequently discovers slices with nontrivial excess calibration error. Quantitatively, raw embeddings produce a positive worst-slice gap in only $58\%$ of model--dataset pairs, whereas the learned calibration geometry does so in $80\%$ of pairs. The paired-difference panel shows that the advantage is heterogeneous across settings: learned $\phi$ discovers larger gaps for many model--dataset pairs. This is consistent with the view that frozen LLM embeddings contain some calibration-relevant information, but in their raw format they are not generally sufficient to expose the full signed heterogeneity of calibration behaviour.

Overall, this ablation shows that the learned representation is not merely applying kernel smoothing in an arbitrary or already sufficient semantic space. Raw hidden-state smoothing mostly recovers a low-variance global overconfidence signal, whereas the learned calibration geometry exposes higher-variance signed structure and more frequently identifies hidden calibration regimes. This supports the role of learning $\phi$: it induces a geometry in which neighbourhoods are organised by calibration-relevant residual structure rather than by the frozen model representation alone.

\subsection{Robustness and Reliability Audits}
\label{app:exp_real:robust}

The main experiments show that the learned miscalibration field
$\widehat{\delta}_\phi$ reveals hidden calibration regimes across datasets
and models. We perform two additional reliability audits to verify that
these regimes are not artifacts of residual noise or optimization
randomness. First, we compare against a permutation null that breaks the
relationship between inputs and residuals. Second, we measure stability
across independent random seeds. Due to computational costs, we run the audits on six 
representative dataset--model pairs with previously identified high or moderate miscalibration heterogeneity.

\paragraph{Permutation-null reliability audit.}
To test whether the discovered structure could arise from finite-sample
noise alone, we construct a permutation null by randomly permuting labels
on the training and validation splits before learning the calibration
geometry. This breaks the systematic relationship between inputs,
confidences, and outcomes while preserving the marginal label distribution
within each split. This audit is especially important because the
worst-slice gap is an adaptive statistic: the slice being evaluated is
chosen after learning $\widehat{\delta}_\phi$, rather than fixed in
advance. As a result, large gaps can arise from searching over many
possible slices even in the absence of true input-dependent calibration
structure. Therefore, for each audited dataset--model pair we run $20$
independent null permutations and rerun the full discovery pipeline,
including representation learning, model selection, and slice selection.
Test labels are left unpermuted and used only for final evaluation. We
then compare the real learned field against the distribution of null
fields.

\begin{table}[h]
    \centering
    \small
    \caption{
\textit{Permutation-null audit of discovered calibration regimes.}
We report the mean and standard deviation over $20$ independent null
permutations. Real fields consistently exhibit larger variability and larger
worst-slice calibration gaps than the null, indicating that the discovered
regimes reflect genuine input-dependent miscalibration rather than residual
noise.
}
    \label{tab:permutation_null}
    \begin{tabular}{llcccc}
    \toprule
    Dataset & Model
    & Std. $\widehat{\delta}_\phi$
    & Null Std. $\widehat{\delta}_\phi$
    & Gap
    & Null Gap \\
    \midrule
    HH-RLHF & Ministral-3-14B
    & 0.173 & $0.052 \pm 0.004$ & 0.117 & $0.000 \pm 0.000$ \\
    HH-RLHF & Ministral-3-3B
    & 0.126 & $0.071 \pm 0.005$ & 0.049 & $0.000 \pm 0.000$ \\
    HH-RLHF & Qwen3-1.7B
    & 0.154 & $0.092 \pm 0.005$ & 0.083 & $0.040 \pm 0.003$ \\
    HH-RLHF & Qwen3-8B
    & 0.199 & $0.174 \pm 0.008$ & 0.135 & $0.063 \pm 0.007$ \\
    MMLU & Llama-3.2-3B
    & 0.170 & $0.151 \pm 0.007$ & 0.124 & $0.025 \pm 0.004$ \\
    MedMCQA & Llama-3.2-1B
    & 0.137 & $0.028 \pm 0.008$ & 0.084 & $0.000 \pm 0.001$ \\
    \bottomrule
    \end{tabular}
\end{table}

Table~\ref{tab:permutation_null} shows that the real learned fields have
larger standard deviation than the corresponding null fields across all
audited dataset--model pairs. More importantly, the real fields identify
regions with substantially larger worst-slice calibration gaps. This second
quantity is the more direct validation of discovery: it shows that the
regions selected by $\widehat{\delta}_\phi$ correspond to genuinely worse
held-out calibration.

The permutation null occasionally produces non-negligible field variation,
for example for Qwen3-8B on HH-RLHF and Llama-3.2-3B on MMLU. This is
expected -- local averaging can amplify random residual fluctuations in
finite samples. However, these null fields do not identify meaningful regions with
high calibration error. In such cases, the real worst-slice gap remains
substantially larger than the null gap. 

The distinction between null runs and the real ones is visible in the training dynamics:
in real runs, the validation proxy continues to improve as the learned
geometry extracts residual structure that transfers to held-out validation
examples. Under label permutation, the validation proxy instead oscillates
near its initial value and early stopping is triggered substantially sooner,
consistent with the absence of learnable input-dependent miscalibration
structure.

Altogether, the audit supports the view
that the learned geometry captures input-dependent calibration structure
rather than simply fitting residual noise.

\paragraph{Seed stability.}
We next evaluate whether the discovered regimes are reproducible across
optimization randomness. For each audited dataset--model pair, we rerun the
full discovery pipeline across multiple random seeds and compare the learned
fields on the held-out test set. We report pairwise field correlation, sign
agreement of $\widehat{\delta}_\phi$, the induced worst-slice calibration
gap, and the standard deviation of the learned field.

\begin{table}[h]
    \centering
    \small
    \caption{
    \textit{Seed stability of the learned miscalibration fields.}
    The fields are highly correlated across
    seeds and yield stable discovered regimes, indicating that the observed
    calibration structure is not an artifact of random initialization.
    }
    \label{tab:seed_stability}
    \begin{tabular}{llcccc}
    \toprule
    Dataset & Model
    & Field corr.
    & Sign agreement
    & Worst-slice gap
    & Std. $\widehat{\delta}_\phi$ \\
    \midrule
    HH-RLHF & Ministral-3-14B
    & $0.87 \pm 0.03$ & $0.79 \pm 0.06$ & $0.13 \pm 0.05$ & $0.17 \pm 0.01$ \\
    HH-RLHF & Ministral-3-3B
    & $0.84 \pm 0.03$ & $0.86 \pm 0.02$ & $0.07 \pm 0.01$ & $0.14 \pm 0.01$ \\
    HH-RLHF & Qwen3-1.7B
    & $0.95 \pm 0.02$ & $0.86 \pm 0.04$ & $0.09 \pm 0.01$ & $0.15 \pm 0.01$ \\
    HH-RLHF & Qwen3-8B
    & $0.96 \pm 0.02$ & $0.83 \pm 0.05$ & $0.13 \pm 0.01$ & $0.20 \pm 0.01$ \\
    MMLU & Llama-3.2-3B
    & $0.98 \pm 0.00$ & $0.91 \pm 0.01$ & $0.13 \pm 0.00$ & $0.17 \pm 0.01$ \\
    MedMCQA & Llama-3.2-1B
    & $0.81 \pm 0.03$ & $0.82 \pm 0.02$ & $0.08 \pm 0.01$ & $0.14 \pm 0.01$ \\
    \bottomrule
    \end{tabular}
\end{table}

Table~\ref{tab:seed_stability} shows that the learned fields are stable
across random seeds. Pairwise field correlations range from approximately
$0.81$ to $0.98$, and sign agreement ranges from approximately $0.79$ to
$0.91$. This indicates that independently trained geometries assign similar
relative miscalibration values and recover similar over- and underconfident
regions. The resulting worst-slice gaps and field standard deviations are
also stable, showing that the discovered regimes have reproducible
calibration consequences.

Taken together, the permutation-null and seed-stability audits provide two
complementary reliability checks. The permutation null shows that the
observed regimes are not explained by random residual fluctuations, while
the seed analysis shows that they are reproducible across independent
training runs. These results support the interpretation that
$\widehat{\delta}_\phi$ captures stable input-dependent calibration
structure.

\subsection{Sensitivity to Regime Definitions}
\label{app:threshold_sensitivity}

The main experiments define discovered over- and underconfident regions using
a fixed reporting threshold $\epsilon=0.05$ on $\widehat{\delta}_\phi$. This
threshold is not tuned separately for each dataset or model; it is used only to
separate examples with clearly signed estimated miscalibration from those with
near-zero values. We therefore evaluate whether the conclusions are sensitive
to this choice.

As an additional check, we also report Brier-score gaps on the same discovered
slices. Since smECE and Brier score are on
different scales, we use smECE only to define non-trivial hidden calibration
structure and report Brier gaps as a complementary metric on the same regions.

For each threshold $\epsilon\in\{0.01,0.05,0.10,0.15\}$, we define
overconfident and underconfident regions using
$\widehat{\delta}_\phi(x)<-\epsilon$ and
$\widehat{\delta}_\phi(x)>\epsilon$, respectively. We then identify the
worst-calibrated discovered slice as the signed region with the larger smECE.
To focus on settings where meaningful hidden calibration structure is present,
Table~\ref{tab:threshold_sensitivity} summarizes dataset--model pairs whose
worst-slice smECE gap exceeds $0.06$ at the corresponding threshold; the
column $N/48$ reports how many of the $48$ dataset--model pairs satisfy this
criterion.

\begin{table}[h]
\centering
\small
\caption{
Sensitivity of discovered regimes to the reporting threshold $\epsilon$.
Rows summarize dataset--model pairs with non-trivial hidden calibration
structure, defined as worst-slice smECE gap greater than $0.06$ at the
corresponding threshold. Brier gaps are computed on the same slice selected by
worst-slice smECE. The column $N/48$ reports the number of such
dataset--model pairs out of $48$ total pairs. Slice size is reported as the
median and interquartile range (IQR) percentage of test examples in the worst
slice.
}
\label{tab:threshold_sensitivity}
\begin{tabular}{lcccccc}
\toprule
$\epsilon$
& Median smECE gap
& Median Brier gap
& \multicolumn{3}{c}{Worst-slice size (\%)}
& $N/48$ \\
\cmidrule(lr){4-6}
& & & Q1 & Median & Q3 & \\
\midrule
0.01 & 0.083 & 0.013 & 20.3 & 55.8 & 64.6 & 14/48 \\
0.05 & 0.083 & 0.021 & 12.5 & 49.8 & 62.5 & 18/48 \\
0.10 & 0.092 & 0.017 & 26.6 & 49.4 & 59.3 & 20/48 \\
0.15 & 0.111 & 0.025 & 24.3 & 47.8 & 55.9 & 18/48 \\
\bottomrule
\end{tabular}
\end{table}

The results show that the main conclusions are not driven by the particular
choice $\epsilon=0.05$. Across all thresholds considered, the median
worst-slice smECE gap remains substantial, ranging from $0.083$ to $0.111$.
The discovered worst slices also remain large: the median worst-slice size is
between $47.8\%$ and $55.8\%$ of the test set. Thus, the discovered regimes are
not small outlier subsets produced by an aggressive threshold. The positive
median Brier gaps across thresholds further indicate that these regions also
correspond to degraded probabilistic prediction quality, not only to artifacts
of the smECE metric.

Finally, we find that the relationship between smECE- and Brier-based
diagnostics is positive. Across dataset--model--threshold
combinations, the Brier gap on the smECE-selected worst slice is positively
correlated with the smECE gap on that slice ($\rho=0.26$). 
Metric-specific worst-slice statistics are more strongly aligned: smECE gaps on smECE- and
Brier-selected slices correlate at $\rho=0.70$, and Brier gaps on the
corresponding slices correlate at $\rho=0.77$. 

Overall, these analyses support the robustness of the discovered regimes to
reasonable choices of the reporting threshold. The same qualitative pattern
persists across thresholds: when hidden calibration structure is present, the
worst discovered slice has a substantial smECE gap, contains a sizeable
fraction of the test set, and also exhibits a positive Brier-score gap. This
suggests that the reported regimes reflect stable input-dependent differences
in probabilistic prediction quality rather than consequences of a particular
threshold choice or calibration metric.

\subsection{Confidence Intervals for Slice-Level Metrics}
\label{app:bootstrap_ci}

To quantify finite-sample uncertainty in the held-out slice-level metrics, we
compute nonparametric bootstrap confidence intervals on the test set. The
learned field $\widehat{\delta}_\phi$ and the induced slice assignments are
kept fixed. For each bootstrap replicate, we resample held-out examples with
replacement and recompute the global smECE, the smECE of the fixed
worst-calibrated slice, and their difference. For slice-size intervals, we
resample the full test set and recompute the fraction of examples assigned to
the fixed worst slice.

These intervals quantify uncertainty in the held-out evaluation conditional on
the learned field and discovered regimes. They do not account for uncertainty
from learning $\phi$ or from adaptive slice selection; those sources of
variation are addressed separately by the seed-stability and label-permutation
audits in Appendix~\ref{app:exp_real:robust}.

Table~\ref{tab:bootstrap_worst_slice} reports bootstrap confidence intervals
under the main reporting threshold $\epsilon=0.05$. Many high-gap settings
have intervals bounded away from zero, indicating that the discovered regimes
correspond to stable held-out calibration differences. Conversely, settings
with near-zero estimated gaps typically have intervals that include zero,
consistent with the absence of substantial hidden calibration structure. The
corresponding slice-size intervals show that large gaps often occur on
non-negligible fractions of the test set rather than isolated examples.

\begin{table}[]
    \centering
    \small
    \caption{
    Bootstrap confidence intervals for discovered worst slices under the main
    reporting threshold $\epsilon=0.05$. 
    (a) Worst-slice smECE gap, defined as the smECE of the worst-calibrated
    discovered slice minus the global smECE on the full test set.
    (b) Percentage of test examples assigned to the same worst-smECE slice.
    Each entry reports the bootstrap mean with a $95\%$ percentile confidence
    interval in brackets. The learned field $\widehat{\delta}_\phi$ and slice
    assignments are kept fixed during bootstrapping, so the intervals quantify
    held-out test-set uncertainty conditional on the discovered regimes.
    Dashes indicate cases where no non-trivial signed worst slice is selected
    under the reporting threshold.
    }
    \label{tab:bootstrap_worst_slice}
    \subfloat[Worst-slice smECE gaps]{
    \begin{tabular}{rcccc}
    \toprule
    dataset & HH-RLHF & MMLU & MMLU-Pro & MedMCQA \\
    model &  &  &  &  \\
    \midrule
    Qwen3-1.7B & 0.083 \scriptsize{[0.070, 0.098]} & 0.074 \scriptsize{[0.048, 0.112]} & 0.013 \scriptsize{[0.000, 0.038]} & 0.002 \scriptsize{[0.000, 0.010]} \\
    Qwen3-4B & 0.101 \scriptsize{[0.071, 0.129]} & 0.062 \scriptsize{[0.046, 0.083]} & 0.031 \scriptsize{[0.000, 0.063]} & 0.002 \scriptsize{[0.000, 0.010]} \\
    Qwen3-8B & 0.132 \scriptsize{[0.116, 0.148]} & 0.069 \scriptsize{[0.049, 0.090]} & 0.066 \scriptsize{[0.029, 0.102]} & 0.002 \scriptsize{[0.000, 0.010]} \\
    Ministral-3-3B & 0.060 \scriptsize{[0.046, 0.076]} & 0.033 \scriptsize{[0.022, 0.047]} & 0.026 \scriptsize{[0.000, 0.059]} & 0.048 \scriptsize{[0.037, 0.058]} \\
    Ministral-3-8B & 0.003 \scriptsize{[0.000, 0.014]} & 0.081 \scriptsize{[0.036, 0.133]} & 0.051 \scriptsize{[0.019, 0.089]} & 0.046 \scriptsize{[0.039, 0.054]} \\
    Ministral-3-14B & 0.118 \scriptsize{[0.102, 0.131]} & 0.010 \scriptsize{[0.003, 0.018]} & 0.046 \scriptsize{[0.013, 0.077]} & 0.057 \scriptsize{[0.049, 0.065]} \\
    Llama-3.2-1B & 0.011 \scriptsize{[0.000, 0.026]} & 0.038 \scriptsize{[0.024, 0.051]} & 0.005 \scriptsize{[0.000, 0.025]} & 0.084 \scriptsize{[0.075, 0.095]} \\
    Llama-3.2-3B & 0.106 \scriptsize{[0.089, 0.122]} & 0.124 \scriptsize{[0.113, 0.136]} & 0.015 \scriptsize{[0.000, 0.050]} & 0.064 \scriptsize{[0.052, 0.074]} \\
    Llama-3.1-8B & 0.002 \scriptsize{[0.000, 0.013]} & 0.064 \scriptsize{[0.053, 0.073]} & 0.029 \scriptsize{[0.000, 0.079]} & 0.082 \scriptsize{[0.068, 0.093]} \\
    gemma-3-1b & 0.067 \scriptsize{[0.052, 0.081]} & 0.002 \scriptsize{[0.000, 0.012]} & 0.110 \scriptsize{[0.069, 0.148]} & 0.002 \scriptsize{[0.000, 0.009]} \\
    gemma-3-4b & 0.002 \scriptsize{[0.000, 0.013]} & 0.020 \scriptsize{[0.008, 0.032]} & 0.005 \scriptsize{[0.000, 0.028]} & 0.004 \scriptsize{[0.000, 0.011]} \\
    gemma-3-12b & 0.040 \scriptsize{[0.026, 0.055]} & 0.026 \scriptsize{[0.014, 0.038]} & 0.005 \scriptsize{[0.000, 0.028]} & 0.002 \scriptsize{[0.000, 0.008]} \\
    \bottomrule
    \end{tabular}
    }
    \hfill
    \subfloat[Percentage ($\%$) of test data in the worst slice]{
    \begin{tabular}{rcccc}
    \toprule
    dataset & HH-RLHF & MMLU & MMLU-Pro & MedMCQA \\
    model &  &  &  &  \\
    \midrule
    Qwen3-1.7B & 26.4 \scriptsize{[26.4, 27.0]} & 4.7 \scriptsize{[3.4, 13.3]} & 43.1 \scriptsize{[41.7, 74.8]} & -- \\
    Qwen3-4B & 11.4 \scriptsize{[10.4, 11.6]} & 15.2 \scriptsize{[15.2, 15.2]} & 46.0 \scriptsize{[45.3, 54.4]} & -- \\
    Qwen3-8B & 35.6 \scriptsize{[35.1, 39.1]} & 11.9 \scriptsize{[11.6, 12.9]} & 34.5 \scriptsize{[7.4, 48.4]} & 20.1 \scriptsize{[20.1, 20.1]} \\
    Ministral-3-3B & 33.4 \scriptsize{[33.4, 33.4]} & 23.6 \scriptsize{[9.3, 34.4]} & 31.1 \scriptsize{[9.2, 36.8]} & 35.9 \scriptsize{[21.9, 38.5]} \\
    Ministral-3-8B & -- & 4.9 \scriptsize{[3.2, 20.0]} & 33.2 \scriptsize{[16.2, 38.7]} & 38.7 \scriptsize{[26.7, 41.6]} \\
    Ministral-3-14B & 17.8 \scriptsize{[17.8, 17.8]} & 27.2 \scriptsize{[24.3, 58.6]} & 28.0 \scriptsize{[5.2, 51.6]} & 48.3 \scriptsize{[38.5, 54.7]} \\
    Llama-3.2-1B & 20.8 \scriptsize{[2.4, 21.4]} & 24.3 \scriptsize{[3.8, 36.4]} & -- & 39.2 \scriptsize{[38.3, 42.5]} \\
    Llama-3.2-3B & 48.3 \scriptsize{[47.3, 50.9]} & 41.1 \scriptsize{[35.6, 42.5]} & 19.3 \scriptsize{[4.5, 22.7]} & 42.0 \scriptsize{[38.7, 42.7]} \\
    Llama-3.1-8B & -- & 39.2 \scriptsize{[27.5, 51.1]} & 12.5 \scriptsize{[5.3, 31.0]} & 39.3 \scriptsize{[39.4, 39.4]} \\
    gemma-3-1b & 30.4 \scriptsize{[30.4, 30.4]} & -- & 0.7 \scriptsize{[0.6, 0.6]} & -- \\
    gemma-3-4b & -- & 22.7 \scriptsize{[22.7, 22.7]} & -- & 40.4 \scriptsize{[40.2, 40.2]} \\
    gemma-3-12b & 36.2 \scriptsize{[36.1, 36.1]} & 22.5 \scriptsize{[22.3, 22.3]} & -- & 24.2 \scriptsize{[20.1, 40.0]} \\
    \bottomrule
    \end{tabular}
    }
\end{table}

\subsection{Post-hoc Interpretation of Discovered Regimes}
\label{app:interpretability}

As a lightweight post-hoc analysis, we examine whether the discovered regimes
align with coarse semantic structure in the data. In particular, we compare
raw LLM embeddings $x$ with the learned calibration representation $\phi(x)$
on MMLU-Pro, where each example is associated with a subject category.

\begin{figure}[h]
    \centering
    \subfloat[Raw embeddings $x$]{
    \includegraphics[width=0.9\linewidth]{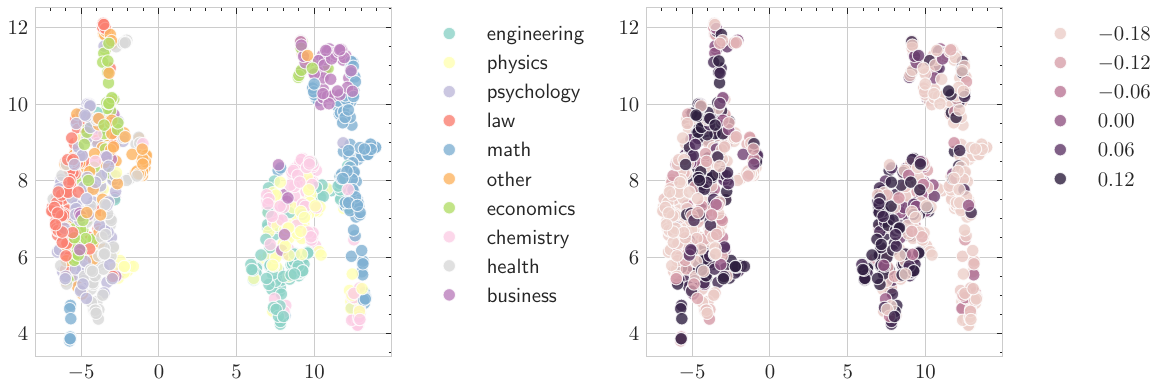}
    }
    \hfill
    \subfloat[Learned embeddings $\phi(x)$]{
    \includegraphics[width=0.9\linewidth]{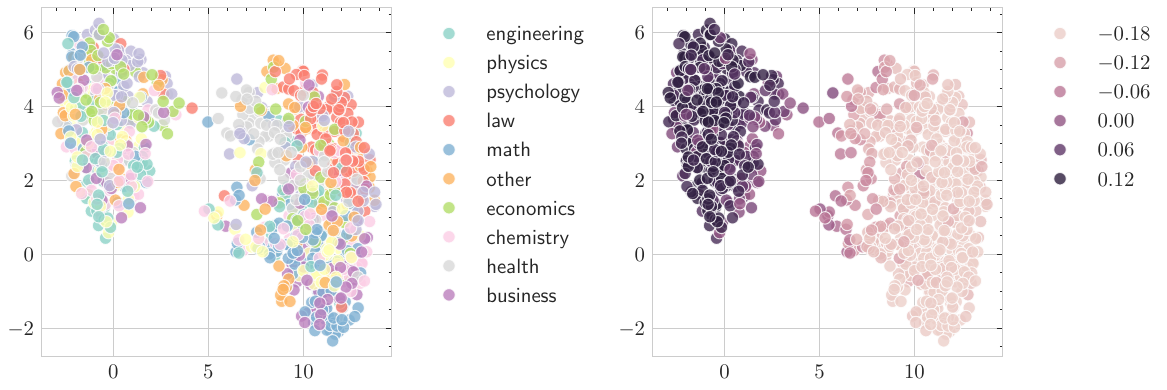}
    }
    \caption{
    \textit{Post-hoc interpretation of the learned calibration geometry on
    MMLU-Pro with Qwen3-8B.}
    (a) UMAP of raw LLM embeddings, colored by subject category (left) and by
    $\widehat{\delta}_\phi$ (right). (b) UMAP of the learned calibration
    representation $\phi(x)$ with the same colorings. Raw embeddings exhibit
    structure partly aligned with subject categories, whereas the learned
    representation separates examples primarily according to the sign and
    magnitude of $\widehat{\delta}_\phi$. Subject categories remain mixed within
    the learned regions, suggesting that the discovered calibration regimes are
    not reducible to coarse topic labels alone.
    }
    \label{fig:mmlu_regime_interpretation}
\end{figure}

Figure~\ref{fig:mmlu_regime_interpretation} shows that the learned
representation substantially reorganizes the input space. While the raw
embeddings exhibit clusters that are largely aligned with subject categories,
the learned representation produces regions that are more clearly separated
by $\widehat{\delta}_\phi$. This indicates that the learned geometry
emphasizes directions predictive of calibration behaviour rather than merely
preserving the semantic organization of the frozen LLM embeddings, reinforcing
the quantitative comparison presented in Appendix~\ref{app:exp_real:raw_vs_learned}.

At the same time, topic structure is not entirely removed: within each
calibration region, examples still show some clustering by subject category.
Thus, the learned geometry appears to separate calibration regimes at a coarse
level while retaining finer semantic organization within them. This suggests a
possible path toward more interpretable regime discovery: future work could
incorporate semantic preservation or disentanglement constraints, or combine
the learned regimes with metadata enrichment and exemplar-based summaries.

\subsection{Detailed Results}\label{app:exp_real:detailed}
\begin{figure}
    \centering
    \includegraphics[width=\linewidth]{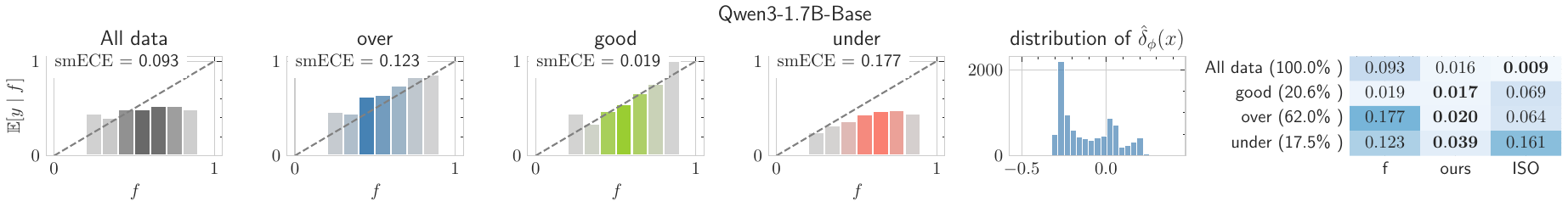}
    \includegraphics[width=\linewidth]{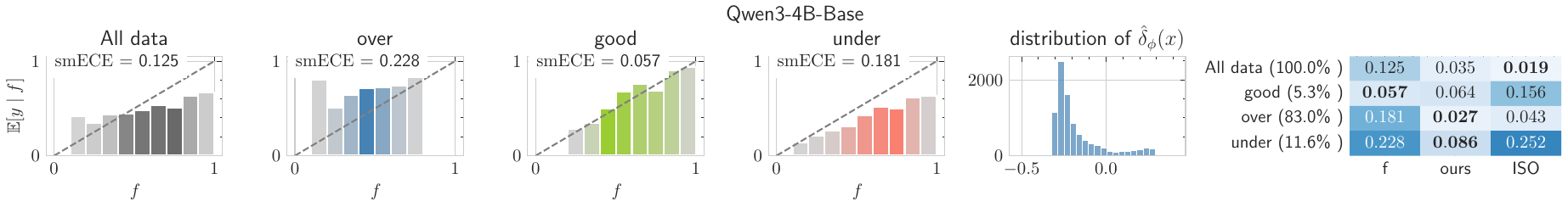}
    \includegraphics[width=\linewidth]{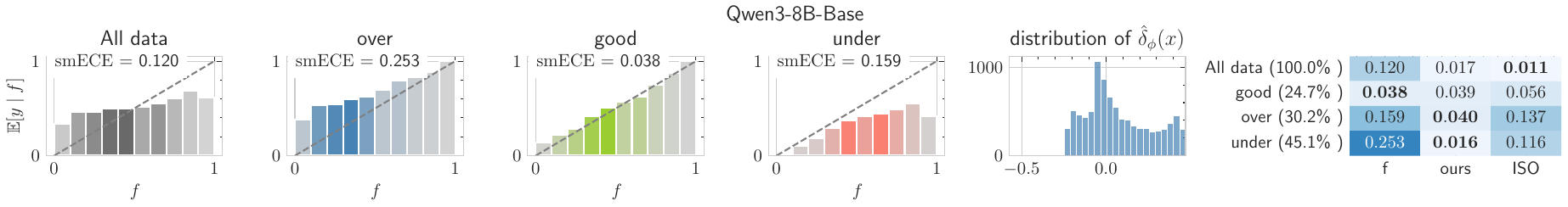}
    \includegraphics[width=\linewidth]{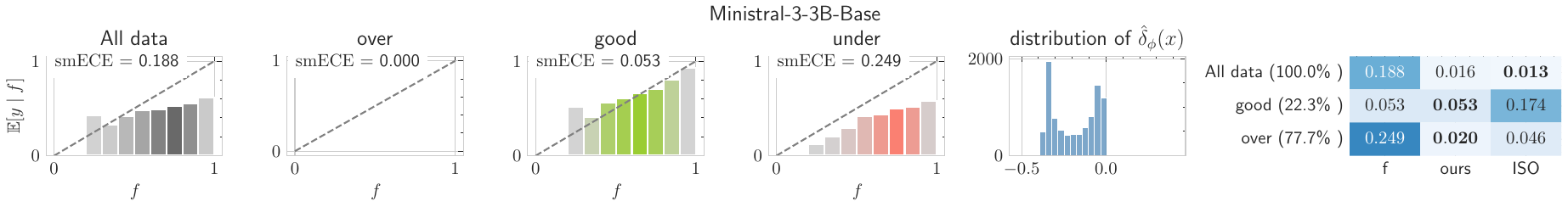}
    \includegraphics[width=\linewidth]{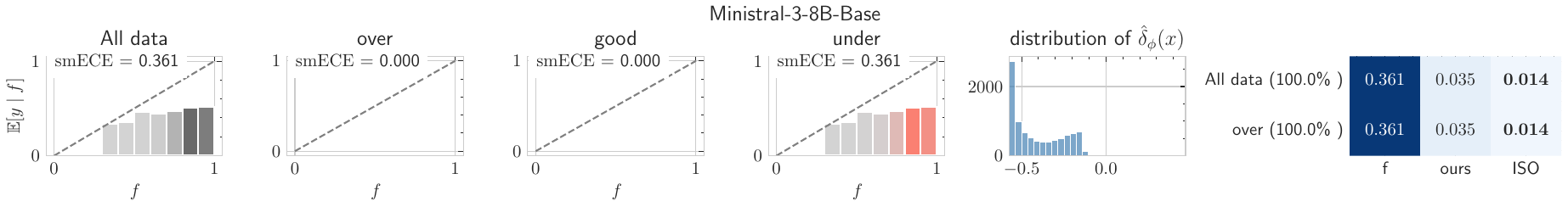}
    \includegraphics[width=\linewidth]{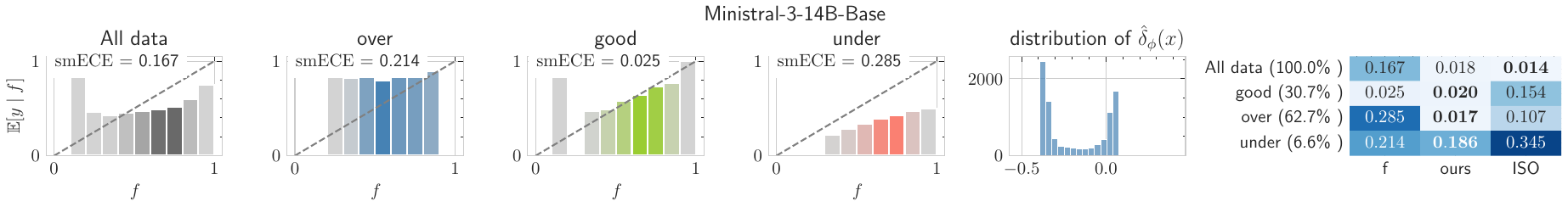}
    \includegraphics[width=\linewidth]{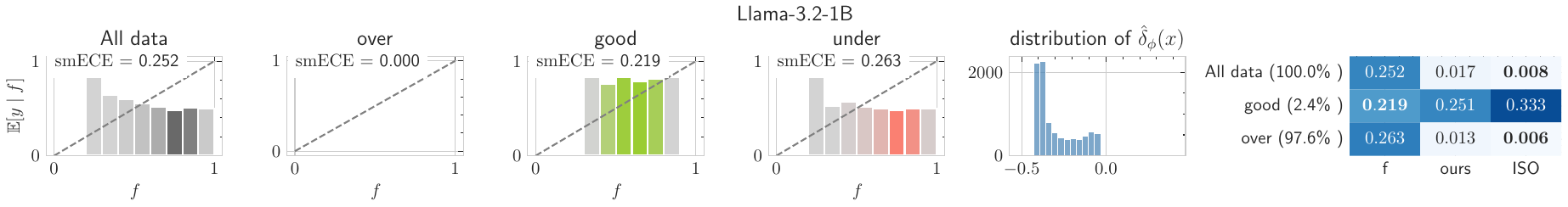}
    \includegraphics[width=\linewidth]{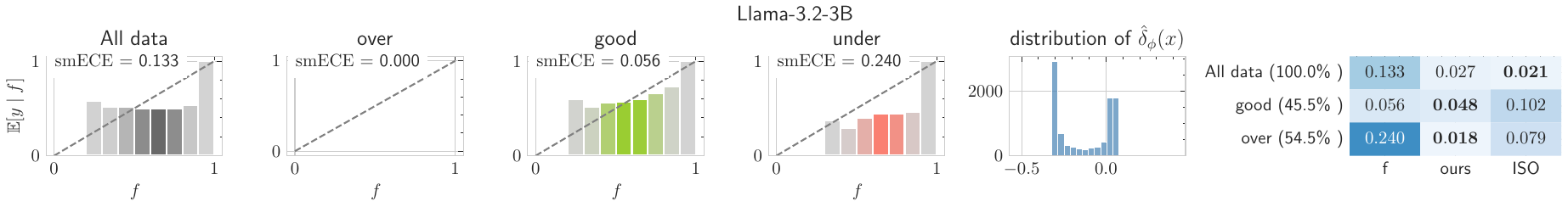}
    \includegraphics[width=\linewidth]{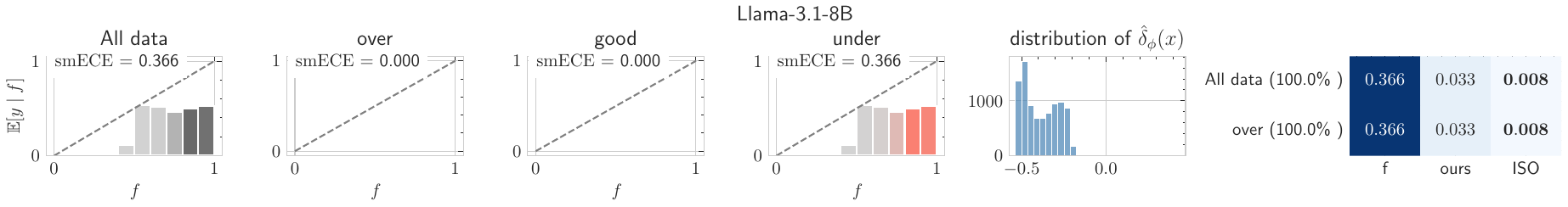}
    \includegraphics[width=\linewidth]{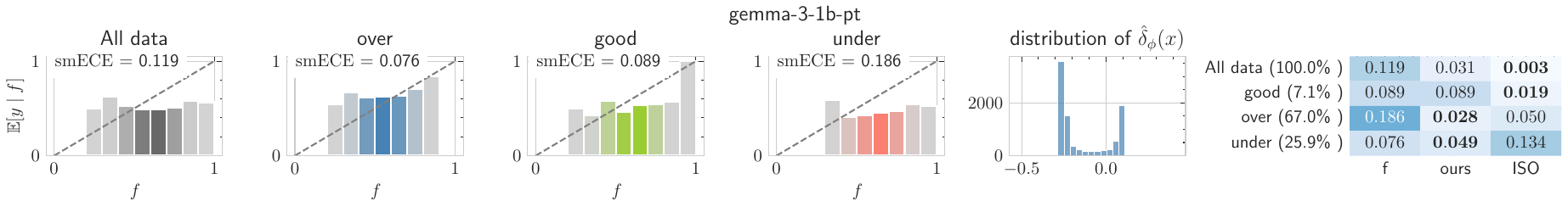}
    \includegraphics[width=\linewidth]{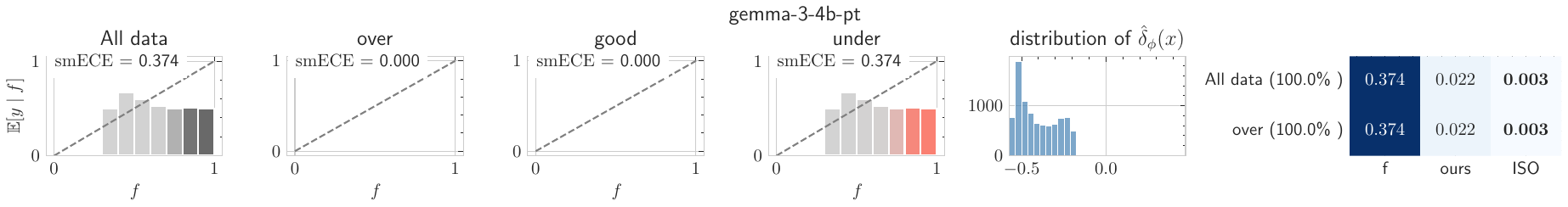}
    \includegraphics[width=\linewidth]{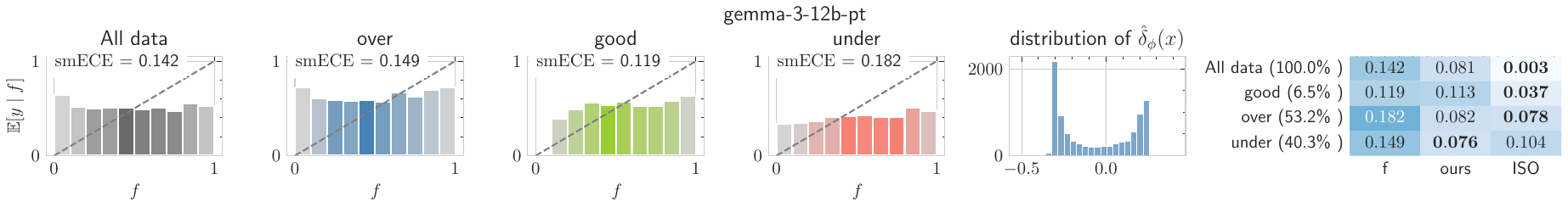}
    \caption{Detailed results for HH-RLHF}
    \label{fig:hhrlhf_detail}
\end{figure}

\begin{figure}
    \centering
    \includegraphics[width=\linewidth]{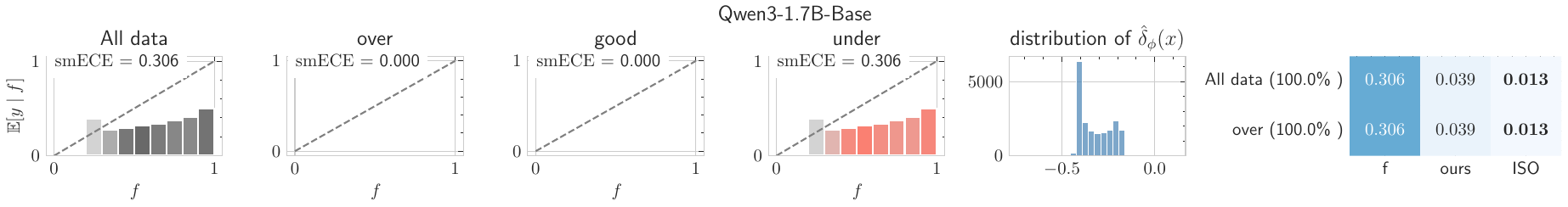}
    \includegraphics[width=\linewidth]{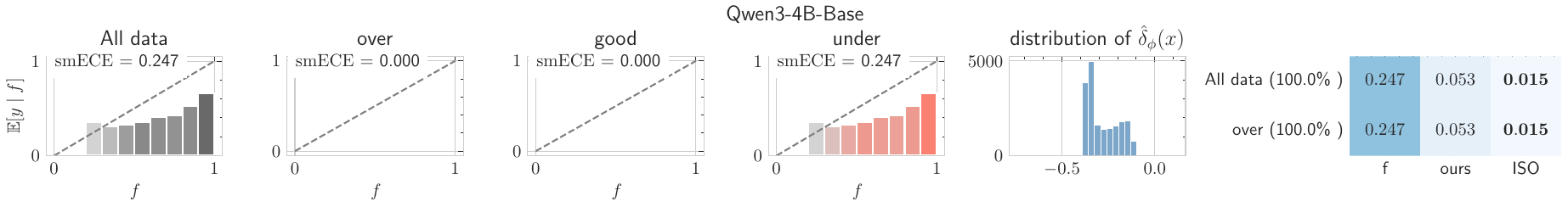}
    \includegraphics[width=\linewidth]{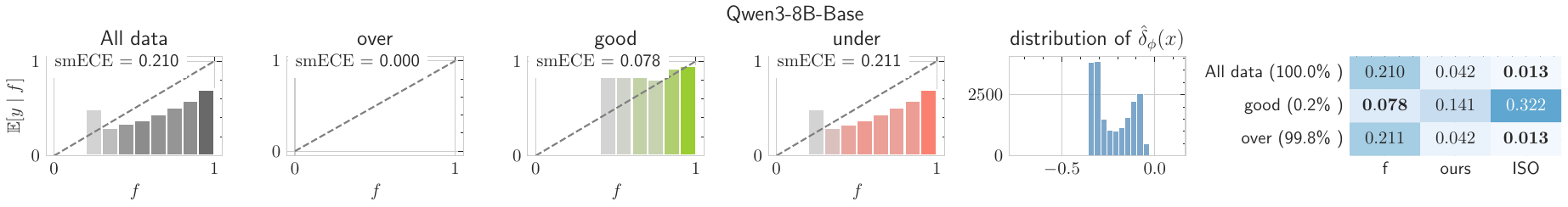}
    \includegraphics[width=\linewidth]{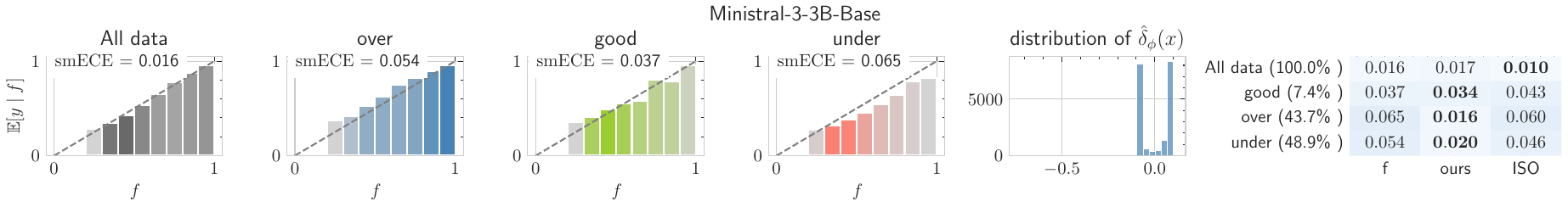}
    \includegraphics[width=\linewidth]{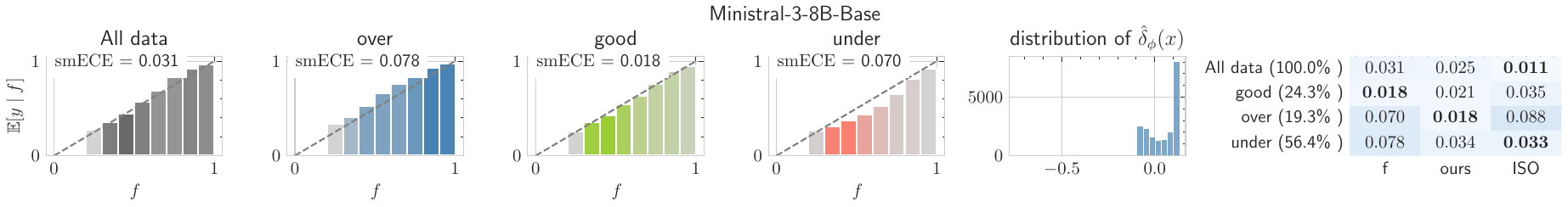}
    \includegraphics[width=\linewidth]{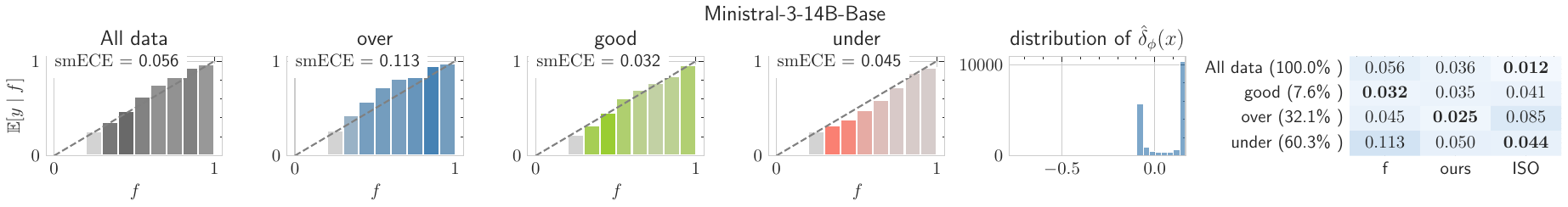}
    \includegraphics[width=\linewidth]{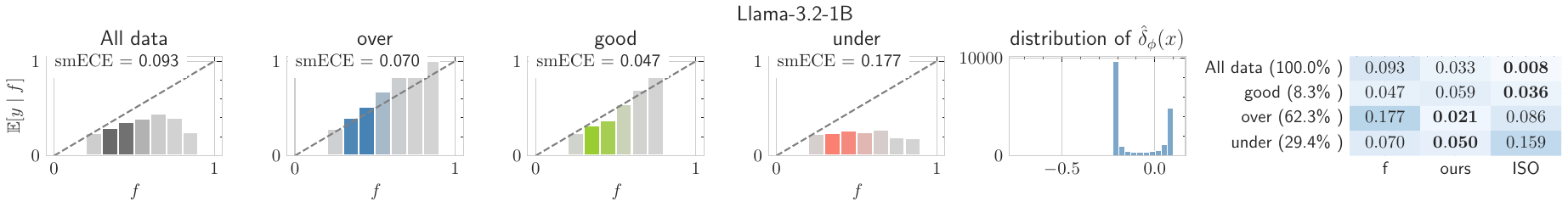}
    \includegraphics[width=\linewidth]{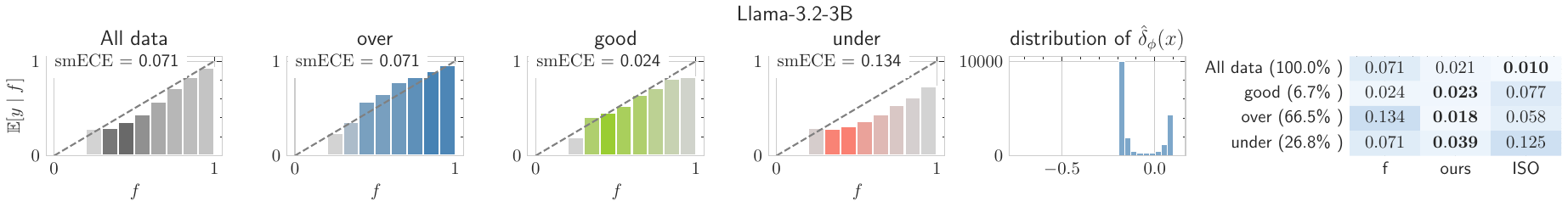}
    \includegraphics[width=\linewidth]{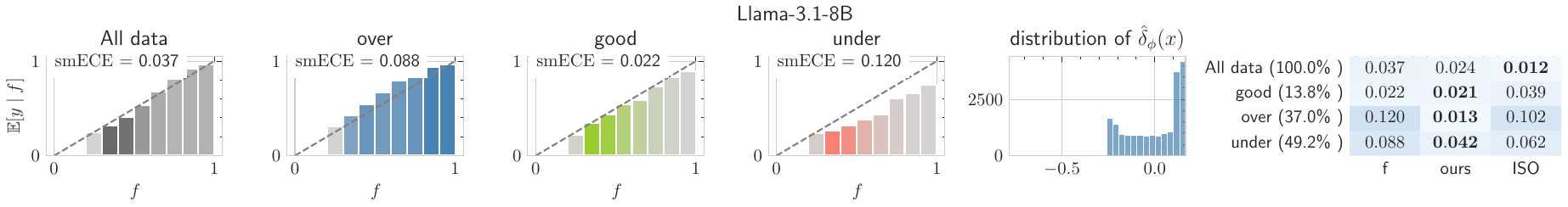}
    \includegraphics[width=\linewidth]{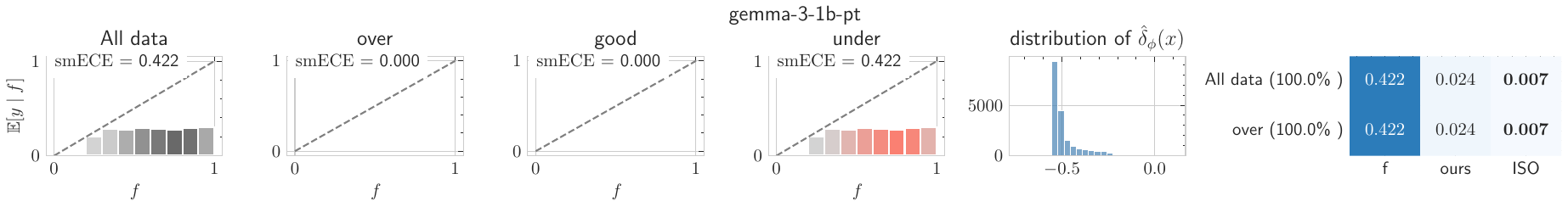}
    \includegraphics[width=\linewidth]{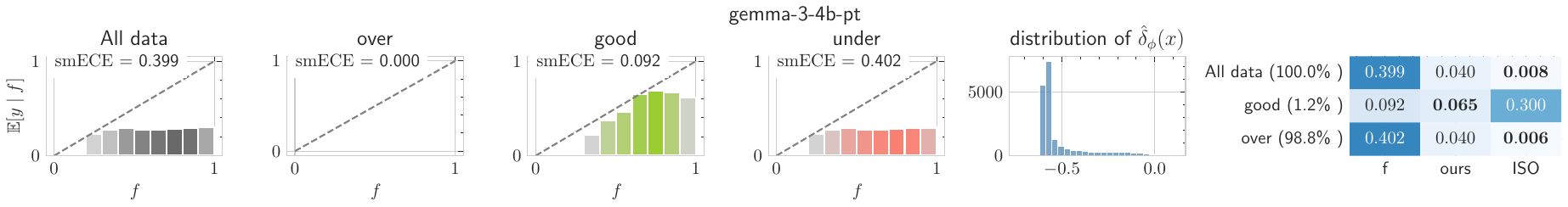}
    \includegraphics[width=\linewidth]{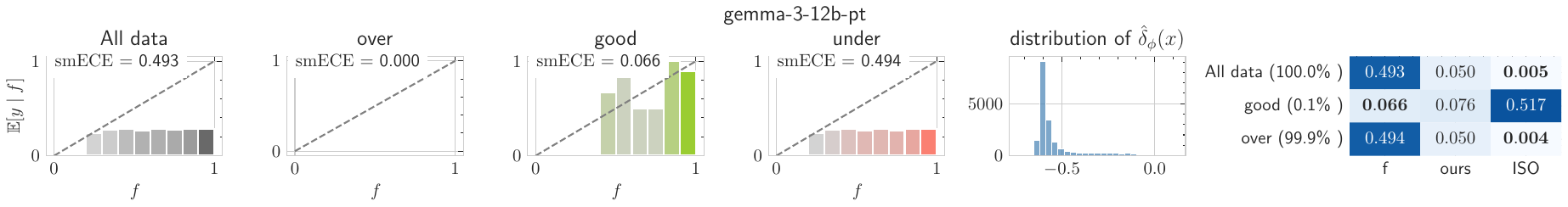}
    \caption{Detailed results for MedMCQA}
    \label{fig:medmcqa_detail}
\end{figure}

\begin{figure}
    \centering
    \includegraphics[width=\linewidth]{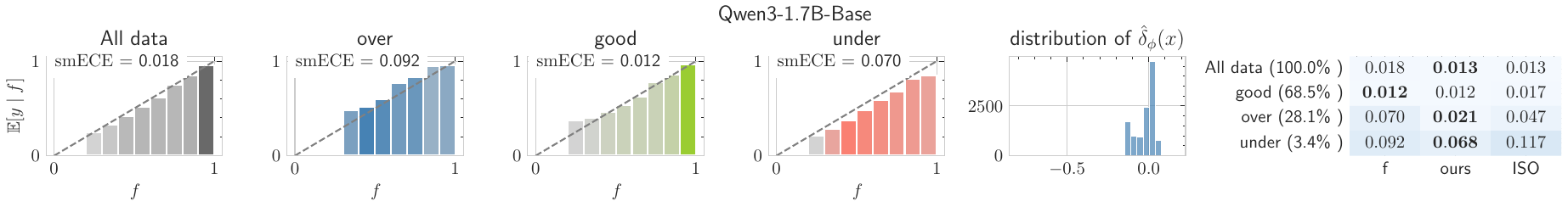}
    \includegraphics[width=\linewidth]{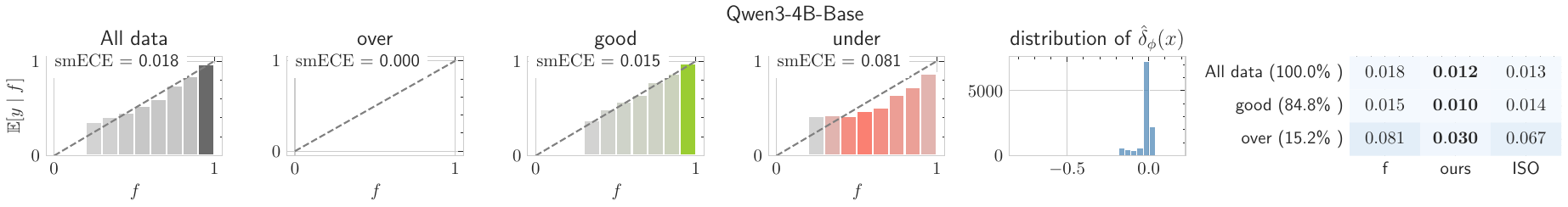}
    \includegraphics[width=\linewidth]{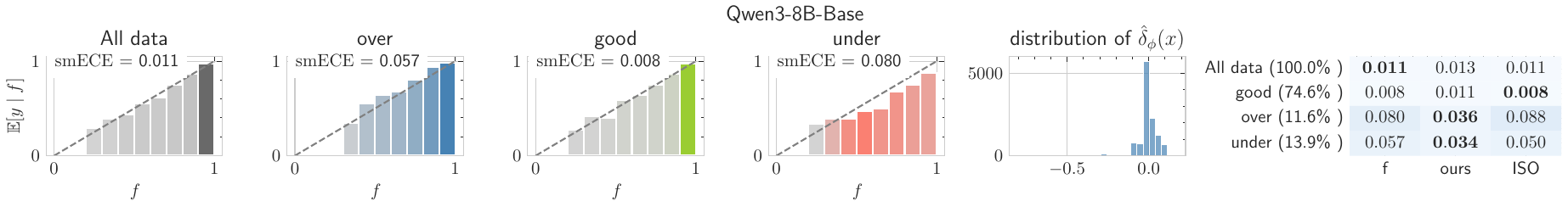}
    \includegraphics[width=\linewidth]{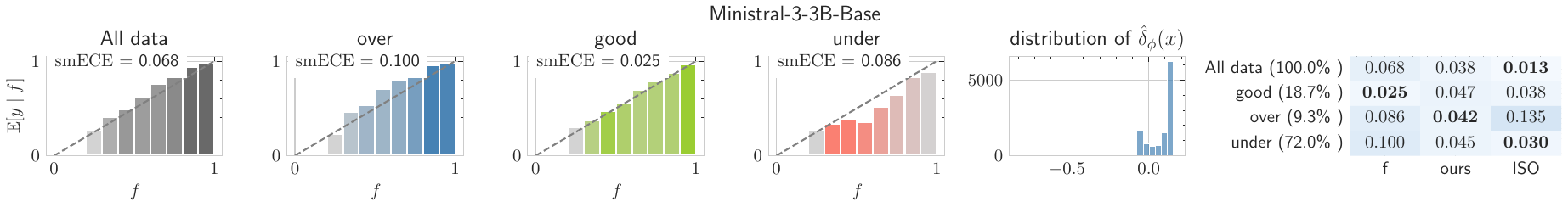}
    \includegraphics[width=\linewidth]{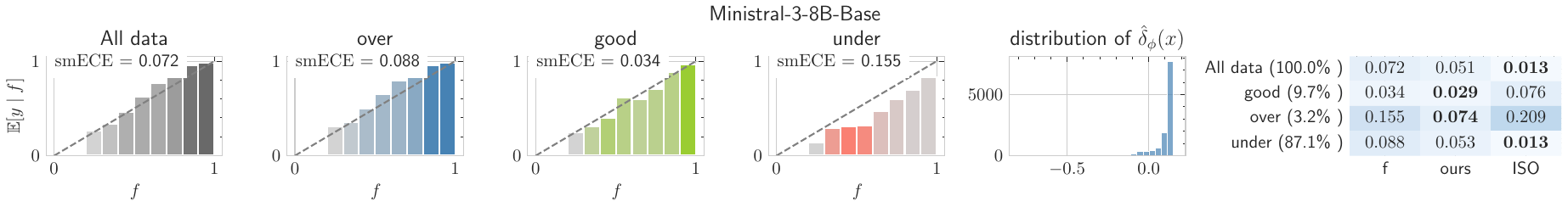}
    \includegraphics[width=\linewidth]{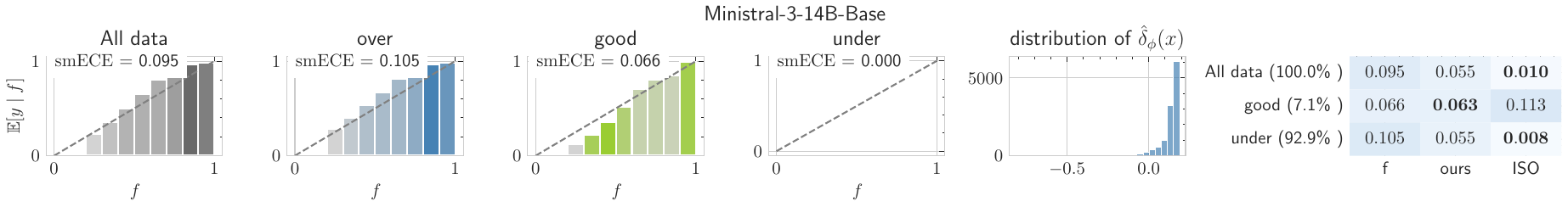}
    \includegraphics[width=\linewidth]{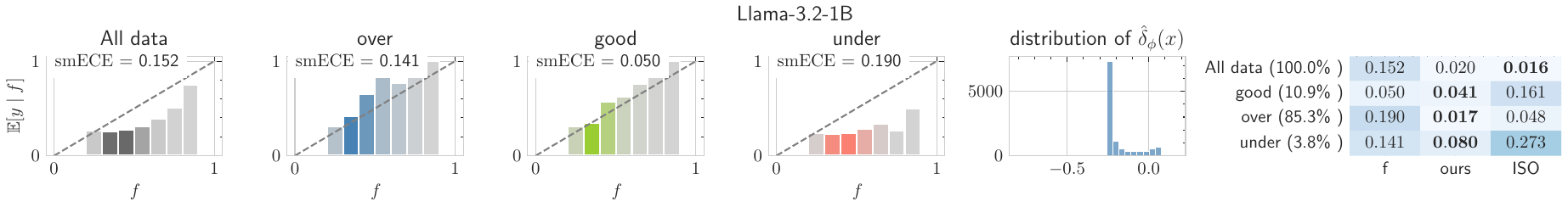}
    \includegraphics[width=\linewidth]{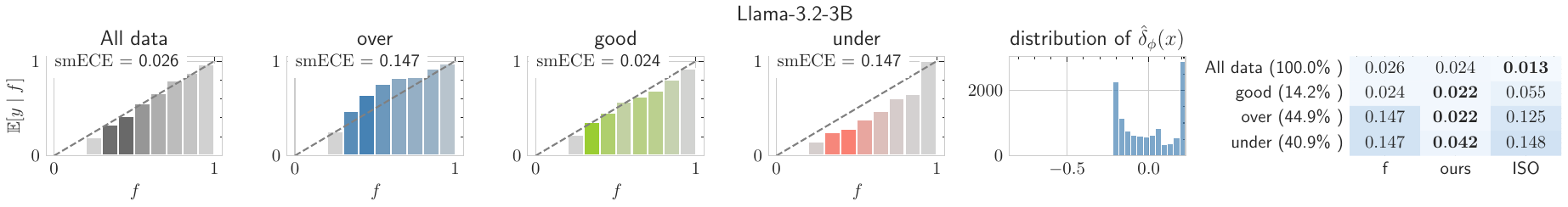}
    \includegraphics[width=\linewidth]{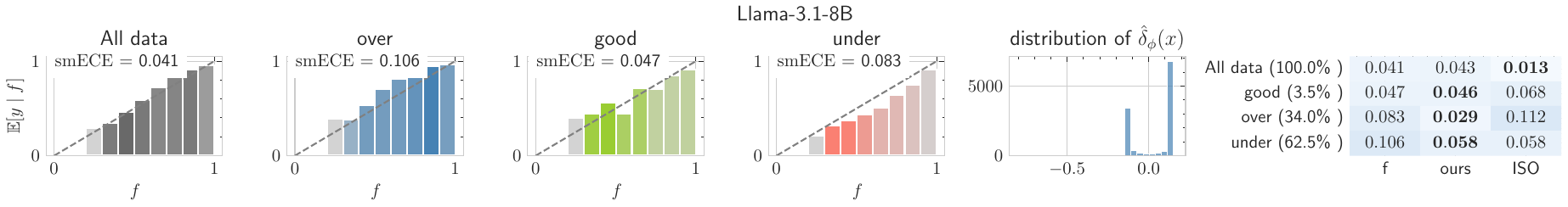}
    \includegraphics[width=\linewidth]{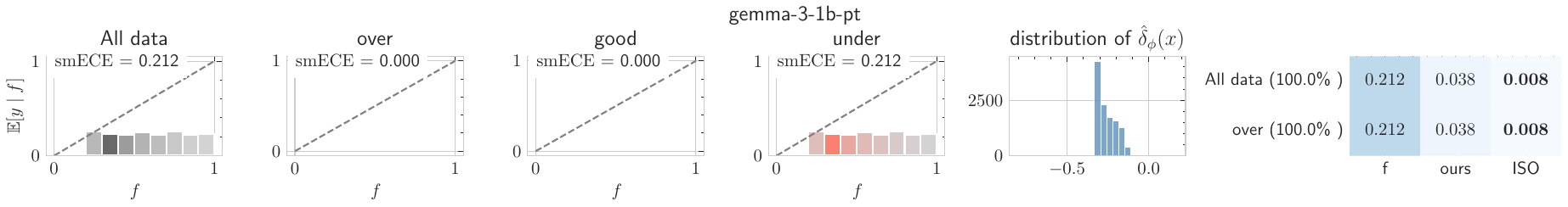}
    \includegraphics[width=\linewidth]{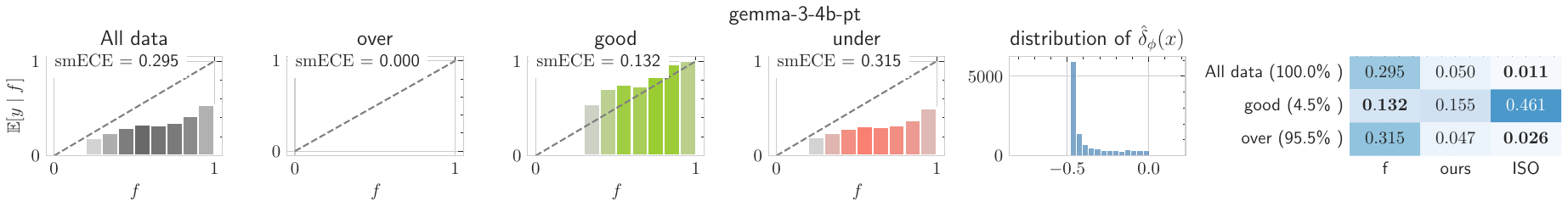}
    \includegraphics[width=\linewidth]{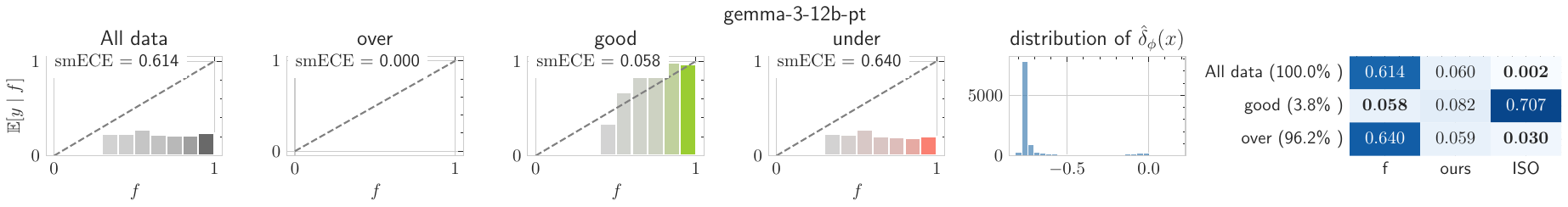}
    \caption{Detailed results for MMLU}
    \label{fig:mmlu_detail}
\end{figure}

\begin{figure}
    \centering
    \includegraphics[width=\linewidth]{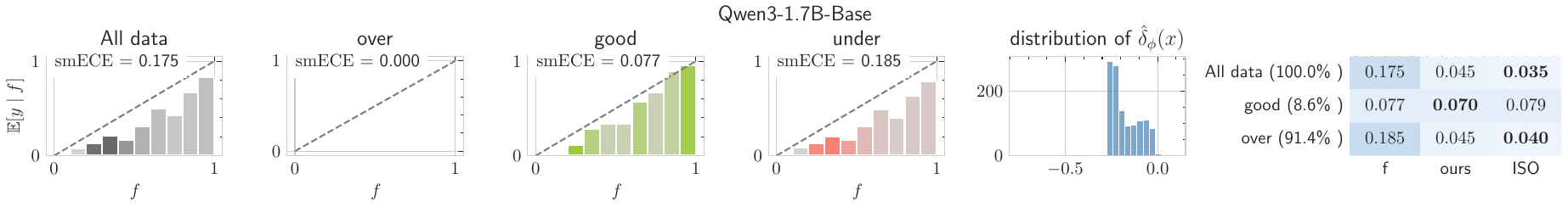}
    \includegraphics[width=\linewidth]{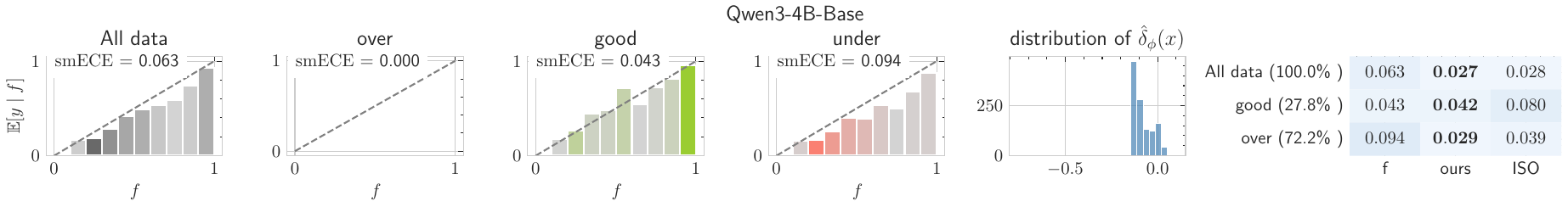}
    \includegraphics[width=\linewidth]{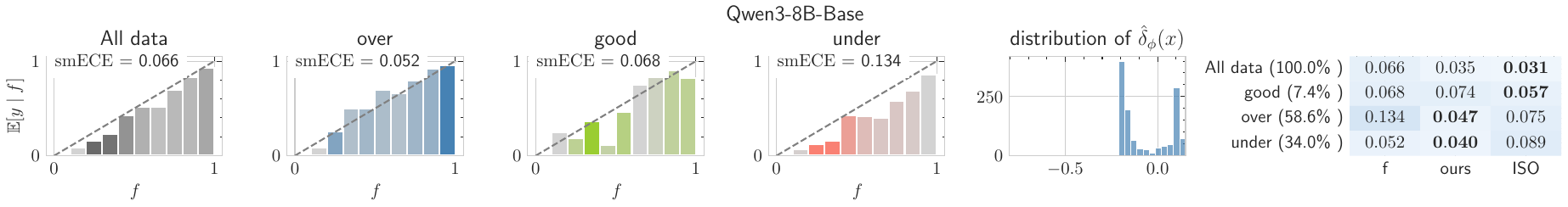}
    \includegraphics[width=\linewidth]{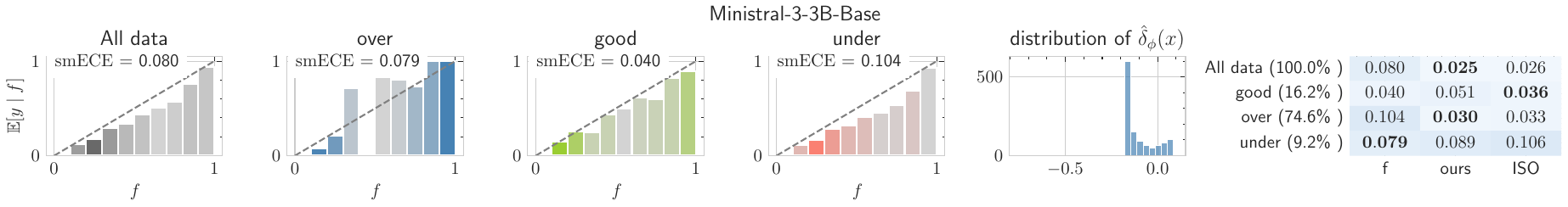}
    \includegraphics[width=\linewidth]{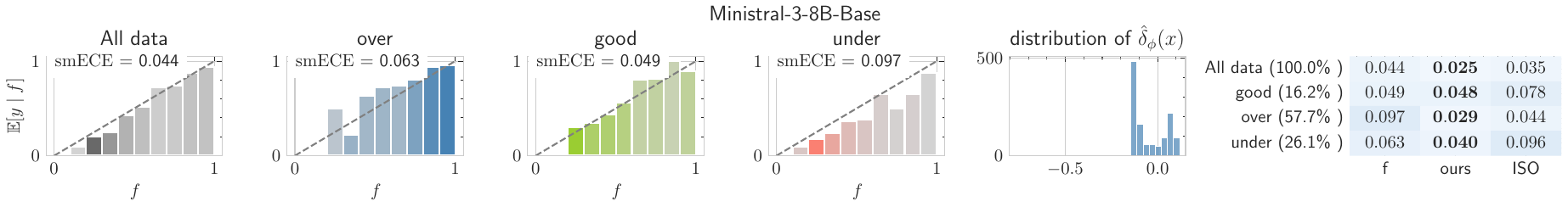}
    \includegraphics[width=\linewidth]{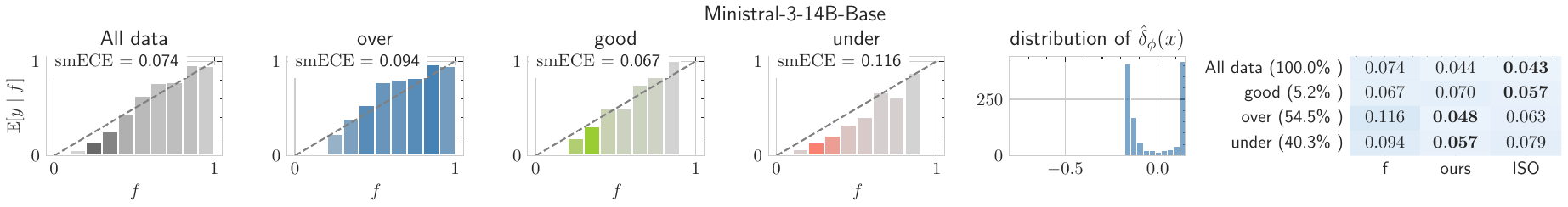}
    \includegraphics[width=\linewidth]{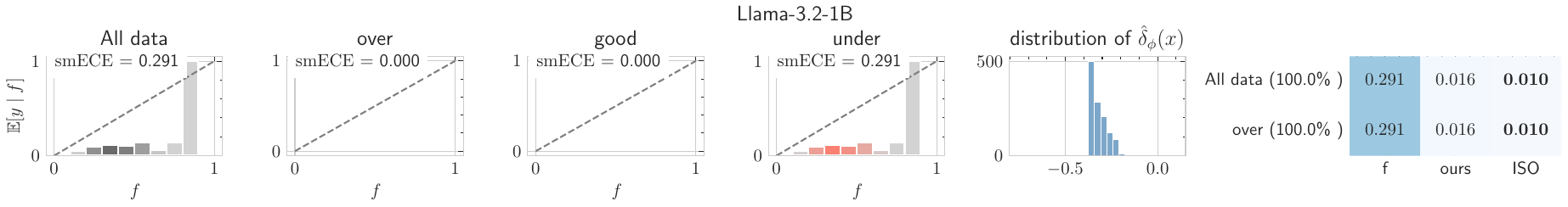}
    \includegraphics[width=\linewidth]{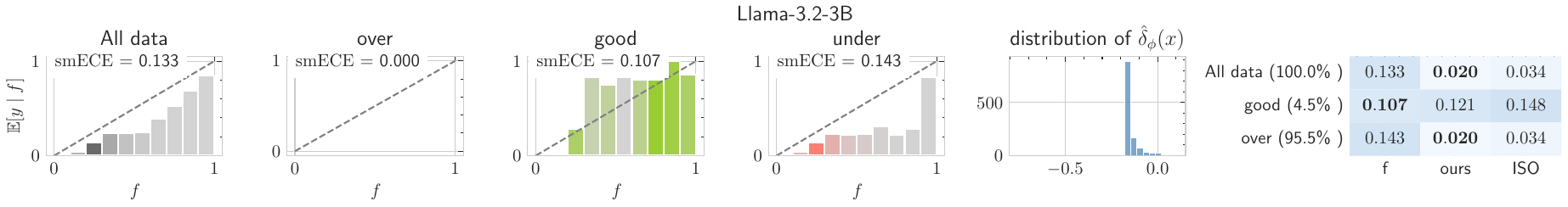}
    \includegraphics[width=\linewidth]{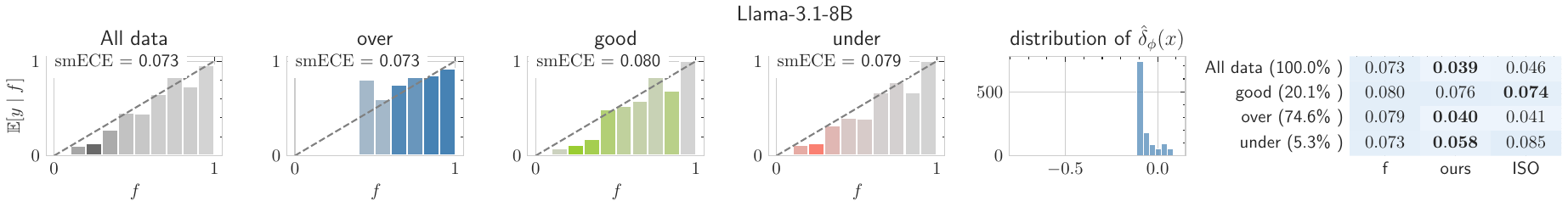}
    \includegraphics[width=\linewidth]{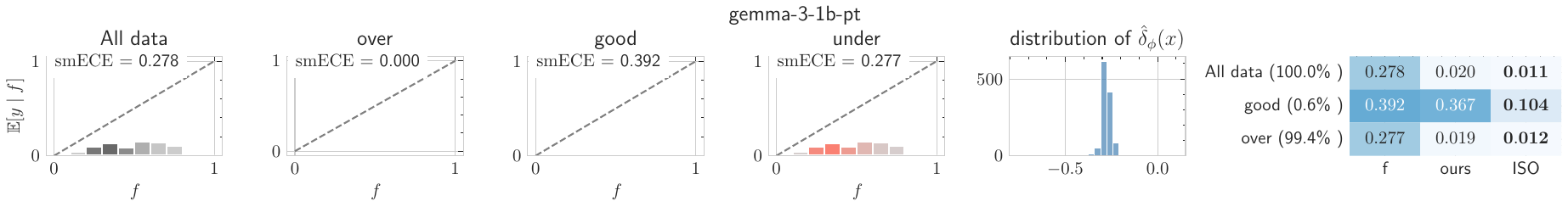}
    \includegraphics[width=\linewidth]{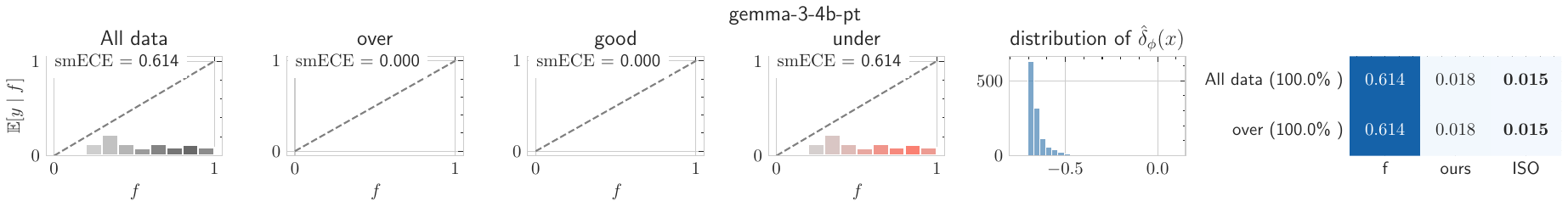}
    \includegraphics[width=\linewidth]{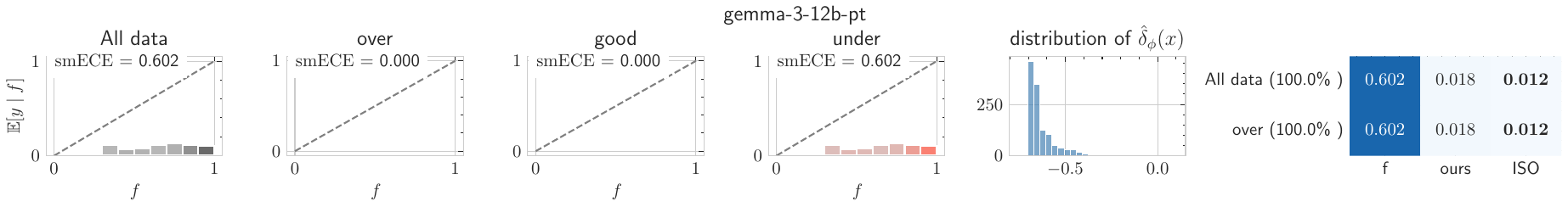}
    \caption{Detailed results for MMLU-Pro}
    \label{fig:mmlu_pro_detail}
\end{figure}

We provide detailed results for each dataset--model pair.
For each pair, we report: (i) the global reliability diagram; (ii)
reliability diagrams for the discovered regions defined by
$\widehat{\delta}_\phi$ using threshold $\epsilon=0.05$; (iii) histograms of
$\widehat{\delta}_\phi$, which show the calibration structure of the learned
field; and (iv) smECE values computed globally and within each
discovered slice. We report smECE for the raw confidence scores $f$, for
our locally corrected scores $\tilde f$, and for isotonic regression (ISO),
which we include as the more flexible confidence-based method compared with
temperature scaling (TS). Results are presented in Figures~\ref{fig:hhrlhf_detail}-\ref{fig:mmlu_pro_detail}

\end{document}